\documentclass[10pt]{iopart}

\newcommand{\uchicagoAA}{Department of Astronomy and Astrophysics, University of Chicago, Chicago, IL 60637}
\newcommand{\kicp}{Kavli Institute for Cosmological Physics, University of Chicago, Chicago, IL 60637}
\newcommand{\fermilab}{Fermi National Accelerator Laboratory, Batavia, IL 60510}

\newcommand{\iaifi}{NSF AI Institute for Artificial Intelligence \& Fundamental Interactions (IAIFI), Cambridge, MA 02139}
\newcommand{\skai}{NSF-Simons AI Institute for the Sky (SkAI), 172 E. Chestnut St., Chicago, IL 60611, USA}

\newcommand{\northeastern}{Department of Physics, Northeastern University, Boston, Massachusetts 02115}
\newcommand{\khoury}{Khoury College of Computer Science, Northeastern University, Boston, Massachusetts 02115}

\newcommand{\githubmaster}{\href{https://github.com/deepskies/SIDDA}{\faGithub}\xspace}
\newcommand{\zenodomaster}{\href{https://zenodo.org/records/15215272}{\faDatabase\xspace}}

\newcommand{\myfigure}{Figure}
\newcommand{\mytable}{Table}

\newcommand{\mysection}{Section}

\newcommand{\myequation}{Equation}

\newcommand{\algoname}{SIDDA}

\expandafter\let\csname equation*\endcsname\relax
\expandafter\let\csname endequation*\endcsname\relax
\usepackage{amsmath}
\usepackage{hyperref}       
\usepackage[capitalize,noabbrev]{cleveref} 

\usepackage{aas_macros}
\usepackage{amsfonts}       
\usepackage{amssymb}
\usepackage{amsthm}
\usepackage{booktabs} 
\usepackage{blindtext}
\usepackage[spanish, english]{babel}
\usepackage[T1]{fontenc}    
\usepackage{fullpage}
\usepackage{graphicx}

\usepackage{algorithmic}
\usepackage{algorithm}

\usepackage{import}
\usepackage[utf8]{inputenc} 
\usepackage{microtype}      
\usepackage{nicefrac}       
\usepackage{subfigure}
\usepackage{textcomp}
\usepackage[textsize=tiny,colorinlistoftodos]{todonotes}
\usepackage{url}            
\usepackage{xargs}    
\usepackage{xcolor}         
\usepackage{array}
\arraybackslash
\usepackage{multirow}
\usepackage{fontawesome}
\usepackage{booktabs}
\usepackage{natbib}
\usepackage{xcolor}
\usepackage{xspace}

\hypersetup{
	colorlinks   = true, 
    urlcolor     = blue, 
    linkcolor    = blue, 
    citecolor    = red   
 }

\begin{document}

\title[...]{\texttt{\algoname{}}: SInkhorn Dynamic Domain Adaptation for Image Classification with Equivariant Neural Networks}

\author{Sneh Pandya\thanks{Corresponding author.}}
\address{\northeastern}
\address{\iaifi}
\address{\fermilab}

\footnotetext{pandya.sne@northeastern.edu}

\author{Purvik Patel}
\address{\khoury}
\author{Brian D. Nord}
\address{\fermilab}
\address{\uchicagoAA}
\address{\kicp}
\author{Mike Walmsley}
\address{Dunlap Institute for Astronomy \& Astrophysics, University of Toronto, Toronto, ON M5S 3H4, Canada}
\address{Jodrell Bank Centre for Astrophysics, Department of Physics \& Astronomy, University of Manchester, Manchester, M13 9PL, UK}
\author{Aleksandra \'{C}iprijanovi\'{c}}
\address{\fermilab}
\address{\uchicagoAA}
\address{\skai}


\begin{abstract} 
Modern neural networks (NNs) often do not generalize well in the presence of a ``covariate shift''; that is, in situations where the training and test data distributions differ, but the conditional distribution of classification labels given the data remains unchanged. In such cases, NN generalization can be reduced to a problem of learning more robust, domain-invariant features.
Domain adaptation (DA) methods include a broad range of techniques aimed at achieving this; however, these methods have struggled with the need for extensive hyperparameter tuning, which then incurs significant computational costs. 
In this work, we introduce \algoname{}, an out-of-the-box DA training algorithm built upon the Sinkhorn divergence, that can achieve effective domain alignment with minimal hyperparameter tuning and computational overhead.
We demonstrate the efficacy of our method on multiple simulated and real datasets of varying complexity, including simple shapes, handwritten digits, real astronomical observations, and remote sensing data. 
These datasets exhibit covariate shifts due to noise, blurring, differences between telescopes, and variations in imaging wavelengths.
\algoname{} is compatible with a variety of NN architectures, and it works particularly well in improving classification accuracy and model calibration when paired with symmetry-aware equivariant neural networks (ENNs). 
We find that \algoname{} consistently enhances the generalization capabilities of NNs, achieving up to a $\approx40\%$ improvement in classification accuracy on unlabeled target data, while also providing a more modest performance gain of $\lesssim 1\%$ on labeled source data.
We also study the efficacy of DA on ENNs with respect to the varying group orders of the dihedral group $D_N$, and find that the model performance improves as the degree of equivariance increases.
Finally, if \algoname{} achieves proper domain alignment, it also enhances model calibration on both source and target data, with the most significant gains in the unlabeled target domain---achieving over an order of magnitude improvement in the expected calibration error and Brier score.
\algoname{}'s versatility across various NN models and datasets, combined with its automated approach to domain alignment, has the potential to significantly advance multi-dataset studies by enabling the development of highly generalizable models.
\githubmaster 
\zenodomaster 
\end{abstract}

\ioptwocol

\section{Introduction}
\label{sec:introduction}

Deep neural networks (NNs) excel at extracting complex features from data, making them a powerful tool for a wide range of tasks, including classification, regression, and anomaly detection.
Unfortunately, some extracted features can be very dataset-specific, which makes it challenging for NN models to generalize to data that differs from the training data, even when the differences are subtle. 
For instance, a significant drop in performance occurs when the input distribution changes between the training and test datasets, despite the conditional distribution of the labels given the inputs remaining the same --- a scenario commonly referred to as a ``covariate shift''~\citep{FS2020,LY2022}. 

Generalization allows models to perform well across diverse data domains, ranging from subtle variations in input distributions to entirely different datasets or environments.  
Differences between training and testing data can be due to data collection or quality \citep{ dodge2016understandingimagequalityaffects}, distortions~\citep{gide2016effectdistortionspredictionvisual}, image corruptions~\citep{ford2019adversarialexamplesnaturalconsequence}, or even single-pixel level differences, which can cause the NN to give inaccurate predictions~\citep{Su_2019}.
Generalization capabilities, in turn, aid the efficiency and applicability of NNs in both science and industry, as they would otherwise need to be continually retrained on new data. 
For example, in astronomy, a generalized model trained on data from one telescope should accurately predict properties of data from another telescope that has different noise characteristics or resolution, significantly accelerating the process of identifying or characterizing celestial objects across surveys.

Domain adaptation (DA) is a group of methods that aim to improve the generalization capabilities of NNs by enabling the NN to learn features in the data that persist across domains~\citep{Wang_2018, wilsongarrett}. 
It is often applied to problems where one has access to labeled data from a ``source'' domain, but would also like the model to perform well on unlabeled ``target'' domain data.
A large group of distance-based DA methods tackles the covariate shift problem by minimizing some distance metric between internal NN latent representations (distributions) of the source and target data. 
This, in turn, forces the NN to extract mainly domain-invariant features, which makes both latent data distributions well-aligned. 

Some well-known distance-based DA methods use maximum mean discrepancy (MMD)~\citep{MMD1}, correlation alignment (CORAL)~\citep{coral}, contrastive domain discrepancy (CDD)~\citep{cdd}, the Kullback-Leibler (KL) divergence~\citep{Kullback_1951}, or the Wasserstein distance~\citep{wd}, which is derived from the optimal transport (OT)~\citep{OT} theory for measuring distance between probability distributions. The theory is aimed at solving the OT problem, which includes determining the minimal cost of transporting a probability mass from one distribution to another, where the cost is defined by a chosen metric that measures the effort required to move it. 
OT thus provides a principled way to quantify the dissimilarity between distributions, capturing both the geometry of the space and the magnitude of the differences. 

In the sciences, NNs have made significant progress --- from mapping the structure of biological proteins \citep{Jumper2021} to the cosmos \citep{jeffrey2024darkenergysurveyyear}. 
In astronomy, astrophysics, and cosmology, the success is mainly due to the emergence of large datasets and high-fidelity simulations.
Stage IV projects, such as the Vera Rubin Observatory Legacy Survey of Space and Time (LSST)~\citep{Ivezi__2019}, the Nancy Grace Roman Telescope~\citep{roman}, and the Euclid mission~\citep{euclid}, will yield an unprecedented amount of data that analysis pipelines must analyze efficiently. 
Simultaneously, there are many simulations of the Universe from sub-parsec to gigaparsec scales with magneto-hydrodynamic simulation suites, such as IllustrisTNG~\citep{nelson2021illustristngsimulationspublicdata} and CAMELS~\citep{Villaescusa_Navarro_2021}, that can be used to prepare pipelines for the analysis of real data. 

Since simulation-trained models often exhibit a substantial drop in performance when applied to real data, there has been extensive work in implementing DA for astronomical applications: for galaxy morphology classification~\citep{deepadversaries, deepastrouda}, classification of supernovae and identification of Mars landforms,~\citep{Vilalta_2019}, inference of cosmology~\citep{Roncoli2023}, inference of strong gravitational lensing parameters~\citep{Swierc2023,Agarwal2024}, constraining star formation histories of galaxies~\citep{Gilda2024}, and gravitational lens finding~\citep{Parul2024}. 
In astronomy, atmospheric distortion, telescope noise, PSF blurring, and data processing errors commonly affect data quality and contribute to the domain shift between simulated and real data.
Most DA applications in astronomy and beyond require extensive hyperparameter tuning that is highly sensitive to the dataset. 
The goal of this work is to address this shortcoming.

Domain shift problems are common in other areas of science and industry, with DA methods being applied to a variety of tasks such as medical image analysis~\citep{Guan2021DomainAF}, classification of remote sensing data~\citep{7486184}, geospatial semantic segmentation~\citep{2023AGUFMIN53A}, autonomous driving in the presence of changing weather conditions~\citep{Li2023DomainAB}, material discovery~\citep{doi:10.1126/sciadv.adr9038}, etc. 
In such domains, shifts can occur across image modalities---for instance, between different wavelengths such as optical and infrared, or, in medical imaging, between MRI and CT scans of the same tissue.

Most image-based problems in the sciences are addressed using various types of convolutional neural networks (CNNs).
At the same time, equivariant neural networks (ENNs) are gaining popularity due to their ability to explicitly encode symmetry information present in the data, which is often explicitly known --- e.g., $\text{SL}(2, \mathbb{C})$ in particle physics \citep{bogatskiy2020lorentzgroupequivariantneural} and $\text{SE}(3)$ for rigid body motion \citep{fuchs2020se3transformers3drototranslationequivariant}. 
ENNs have also been shown to achieve state-of-the-art performance on many tasks, such as 3D object classification and alignment \citep{esteves2018learningso3equivariantrepresentations}, dynamical systems modeling, representation learning in graph autoencoders and predicting molecular properties~\citep{pmlr-v139-satorras21a}, classification and segmentation tasks~\citep{deng2021vectorneuronsgeneralframework}, and accounting for function-preserving scaling symmetries that arise
from activation functions~\citep{kalogeropoulos2024scaleequivariantgraphmetanetworks}.
They have also been shown to possess inherent robustness to symmetric transformations and noise perturbations due to their restricted feature learning \citep{pandya2023e2equivariantneuralnetworks, Bulusu_2022, du2022se3equivariantgraphneural}. 
This robustness has been observed to increase with the group order \( N \) for the cyclic group \( C_N \) and dihedral group \( D_N \), which are subgroups of $\text{SO}(2)$ and $\text{O}(2)$, respectively. 
Still, in the presence of covariate shifts, even ENNs can exhibit a drop in performance~\citep{pandya2023e2equivariantneuralnetworks}.

In this work, we focus on the Sinkhorn divergence \citep{altschuler2018nearlineartimeapproximationalgorithms}, a symmetrized variant of regularized OT distances.
We introduce SInkhorn Dynamic Domain Adaptation (\algoname{}), a more automated training algorithm for DA that minimizes the need for hyperparameter tuning.
To achieve this, we leverage active scaling of (1) the entropic regularization of the OT plan, and (2) the weighting of classification and DA loss (i.e., for addressing the covariate shift in the datasets) terms, during training. 
To demonstrate the efficacy and broad scope of applications of our proposed DA method, in this work, we use several datasets of varying complexity: simple simulated datasets, well-known benchmark datasets used in the computer science community,  real observational galaxy datasets, and real remote sensing datasets.
We also study the robustness of ENNs and the improved efficacy of \algoname{} when used in conjunction with ENNs compared to typical CNNs. 

The paper is organized as follows. 
In Section \ref{sec:methods}, we describe existing DA methods and their shortcomings, motivating the use of OT-based distances to facilitate DA. 
We describe the core methodology of \algoname{}, and motivate the use of ENNs as inherently robust architectures. 
In Section \ref{sec:data}, we describe the construction of all simulated and real datasets we use in our study. 
In Section \ref{sec:experiments}, we describe the network architectures used, training procedures, metrics for calibration, and metrics for interpreting NN latent spaces. 
In Section \ref{sec:results}, we summarize our results and conclude in Section \ref{sec:conclusion}.

\section{Methods}
\label{sec:methods}

The major components of our work are DA and equivariance, which we combine to create a more efficient and robust NN classifier.
Within DA, we implement the Sinkhorn divergence, a symmetrized and regularized variant of OT distances that offers considerable improvement in DA over traditional methods. 
We construct a training program that constantly adjusts the loss landscape and regularization strength of the Sinkhorn plan, offering optimal domain alignment with minimal hyperparameter tuning.

\subsection{Domain Adaptation}
\label{sec:methods:domain_adaptation}

DA comprises a set of techniques aimed at aligning the latent distributions of NNs in the presence of covariate shifts in data. 
Typically, DA operates in settings where one can access labeled source images $x\in \mathbf{X_s} \subseteq \mathbb{R}^{m \times m}$, and unlabeled target images $x^*\in \mathbf{X_t} \subseteq \mathbb{R}^{m \times m}$ from $\mathbf{X}_s$ and $\mathbf{X}_t$ source and target data domains, where $m$ denotes the number of pixels in each dimension (height and width) of the image. 

Consider the latent vectors $z \in \mathbf{Z}_s \subseteq \mathbb{R}^l$ and $z^* \in \mathbf{Z}_t \subseteq \mathbb{R}^l$, where $\mathbf{Z}_s$ and $\mathbf{Z}_t$ denote the latent spaces of the source and target domains, respectively, and $l$ represents the dimension of the latent vectors (i.e., the width of the corresponding neural network layer).
Latent distributions refer to the probability distributions over these latent vectors, and during training, DA minimizes a statistical distance measure between them.
DA is incorporated through an additional loss term, $\mathcal{L}_\text{DA}$, alongside the standard task loss (e.g., cross-entropy for classification, $\mathcal{L}_\text{CE}$), to promote alignment between the two latent distributions.
In this work, $\mathcal{L_{\text{DA}}}$ (the ``DA loss'') is the loss due to covariate shifts.
The total loss function is then:
\begin{equation}
\label{eqn:CEDAloss}
    \mathcal{L} \propto \mathcal{L_{\text{CE}}} + \mathcal{L_{\text{DA}}} \;.
\end{equation}
In practice, a delicate balance between the two terms must be achieved to ensure proper alignment.

There are numerous DA methods, each with its own strengths and limitations. 
One commonly used approach is MMD \citep{MMD1, MMD2}, where the distance between the means of the latent embeddings from the source and target domains serves as the DA loss function. 
In DA, comparisons are often made between distributions that are not explicitly known but can be sampled. 
MMD can be combined with kernel methods, which map probability distributions into a high-dimensional reproducing kernel Hilbert space (RKHS) \citep{MMD2}, providing a more flexible method for comparing distributions. 
This approach allows for analyzing distributions through well-defined operations in the RKHS, even when the original distributions are not well-defined. 
The MMD between two probability distributions $\mu$ and $\nu$ --- representing distributions over latent vectors $z$ and $z^*$, respectively --- is  
\begin{align}
    \text{MMD}(\mu, \nu) = &\Bigg( \mathbb{E}_{z, z' \sim \mu} \left[ k(z, z') \right] + \mathbb{E}_{z^*, z^{*'} \sim \nu} \left[ k(z^*, z^{*'}) \right] \nonumber \\ 
    &- 2 \mathbb{E}_{z \sim \mu, z^* \sim \nu} \left[ k(z, z^*) \right] \Bigg)^{1/2} \; ,
\end{align}
where $k$ represents the kernel function, and $z, z'$, and $z^*$, $z^{*'}$ are individual samples from latent distributions $\mu$ and $\nu$, respectively. 

Despite its utility, MMD has several theoretical and implementation-related shortcomings. 
First, its efficacy is highly sensitive to the choice of $k$. 
In typical applications, the Gaussian kernel $k({z}, {z}^*) = \exp\left(- \frac{|| {z} - {z}^* ||^2}{2 \epsilon^2}\right)$ is used with kernel bandwidth $\epsilon$. 
Other kernel options include the linear kernel $k(z, z^*) = z^T z^*$, the Laplacian kernel $k(z, z^*) = \exp\left(- \frac{|| {z} - {z}^* ||}{2 \epsilon}\right)$, and others. 
Most kernels generally belong to a one-parameter family (e.g., $\epsilon$ for Gaussian and Laplacian kernels) and must be carefully tuned, or complex linear combinations of kernels with many different parameter values must be used.
The specific choice of kernel depends heavily on the nature of the problem. 
That is, MMD can exhibit bias with small sample sizes and often struggles with domain alignment when dealing with high-dimensional distributions \citep{Muandet_2017, reddi2014decreasingpowerkerneldistance}.

\subsection{Optimal Transport and The Sinkhorn Divergence}
\label{sec:methods:sinkhorn}

OT distances and their symmetrized variants, such as Sinkhorn divergences, offer an alternative to MMD. 
Traditionally, computing OT is prohibitively expensive \citep{peyré2020computationaloptimaltransport}.
Entropic regularization, $\text{OT}_\sigma$, \citep{dessein2018regularizedoptimaltransportrot} provides a more efficient method for estimating OT distances. 
The regularized OT is defined as
\begin{equation}
    \text{OT}_{\sigma}(\mu, \nu) = \min_{\gamma \in U(\mu, \nu)} \left( \sum_{i,j} \gamma_{ij} d(z_i, z^*_j)^p + \sigma H(\gamma) \right),
\end{equation}
where $d(z_i, z^*_j)^p$ is the distance between source feature $z_i$ and target feature $z^*_j$.
When $p=1$, this distance becomes the Earth Mover's distance \citep{710701}, and when $p=2$, it becomes the quadratic Wasserstein distance.
The transport plan $\gamma \in U(\mu, \nu)$ is a joint probability distribution between $\mu$ and $\nu$, where the set of admissible transport plans $U(\mu, \nu)$ is defined by the marginal constraints:
\begin{equation}
    \sum_j \gamma_{ij} = \mu_i, \quad \sum_i \gamma_{ij} = \nu_j.
\end{equation}
The entropy $H(\gamma) = - \sum_{i,j} \gamma_{ij} \log \gamma_{ij}$ regularizes the transport plan $\gamma$, and $\sigma$ controls the regularization strength. 
One limitation of $\text{OT}_\sigma$ is that $\text{OT}_{\sigma}(\mu, \mu) \neq 0$, implying a non-zero cost even when transporting a distribution to itself, leading to bias in the measure.

To correct this bias, the Sinkhorn divergence $S_{\sigma}(\mu, \nu)$, defined as 
\begin{equation}
    S_{\sigma}(\mu, \nu) = \text{OT}_{\sigma}(\mu, \nu) - \frac{1}{2} \text{OT}_{\sigma}(\mu, \mu) - \frac{1}{2} \text{OT}_{\sigma}(\nu, \nu),
\end{equation}
can compensate for the bias in $\text{OT}_{\sigma}$ \citep{feydy2018interpolatingoptimaltransportmmd}. 
As $\sigma \to 0$, $S_{\sigma}(\mu, \nu)$ converges to the true optimal transport $\text{OT}_0$, and as $\sigma \to \infty$, it interpolates towards MMD loss \citep{feydy2018interpolatingoptimaltransportmmd}. For small values of $\sigma$, an unbiased transport plan that still enjoys the benefits of OT-based distances can be constructed.

\subsection{Dynamic Sinkhorn Divergences for Domain Adaptation}
\label{sec:methods:dynamic_optimal_transport}

For this work, $\mathcal{L}_\text{DA}$ in \myequation{}~\ref{eqn:CEDAloss} is specifically the Sinkhorn divergence $S_{\sigma}(\mu, \nu)$. 
However, a careful balance between $\mathcal{L}_\text{CE}$ and $\mathcal{L}_\text{DA}$ must be achieved to optimize the classification task while simultaneously maximizing domain alignment. 

Finding the best weights for each of the loss terms can be very challenging and time-consuming. 
Furthermore, a single choice of weights might not be the best choice throughout the whole training procedure. 
To manage this balance, we employ dynamic weighting of the losses by introducing two trainable parameters, $\eta_1$ and $\eta_2$, which 
dynamically adjust the contributions of the loss terms for each task throughout training. 
These parameters ensure that no single loss term dominates the optimization process, allowing the loss landscape to be optimally adjusted for both tasks. 
Drawing inspiration from~\cite{kendall2018multitasklearningusinguncertainty}, we use the following for the total loss function:
\begin{equation}
\label{eqn:DOT}
    \mathcal{L}= \frac{1}{2 \eta_1^2}\mathcal{L_{\text{CE}}} + \frac{1}{2 \eta_2^2} \mathcal{L_{\text{DA}}} + \log(| \eta_1 \eta_2 | ) \;,
\end{equation}
where $\eta_1$ and $\eta_2$ are trainable scalars, and their values are jointly learned with the model weights during training. 
At the beginning of training, both $\eta_1$ and $\eta_2$ are initialized to a value of one and are subsequently updated during training.
The inclusion of the term $\log (|\eta_1 \eta_2|)$ acts as a regularization to prevent $\eta_1$ and $\eta_2$ from collapsing to unstable values, such as zero.
As $\eta_i \rightarrow 0$, the corresponding loss term is more heavily weighted. 
To ensure that no single component dominates, we impose the additional constraint $\eta_2/\eta_1 \geq 0.25$. 
In general, the DA term must not dominate over the classification loss, which the above inequality enforces.
For our implementation, we found that this threshold worked best and stabilized training, but such a cutoff may not always be optimal. 
 
In~\cite{kendall2018multitasklearningusinguncertainty}, the two weight terms  $\eta_1$ and $\eta_2$ were introduced for the dynamic weighting of the losses. 
These terms explicitly minimize the regression uncertainty associated with each loss term, as their model outputs a Gaussian distribution with variance $\eta_i^2$ for each task.
In the case of classification, their weight terms become $1 / \eta_i^2$. 
Since uncertainties are not one of the network outputs in our case, the exact written form of loss weights is not important and the extra factor of two can very well be absorbed into the trainable weight parameter.

The level of regularization $\sigma$ in $S_{\sigma}(\mu, \nu)$ is another critical hyperparameter \citep{feydy2018interpolatingoptimaltransportmmd}.
When $\sigma$ is too small, the transport plan approaches the true Wasserstein distance, substantially increasing the computational cost of domain alignment. 
In this regime, the Sinkhorn iterations may also fail to fully converge, potentially introducing biases similar to those observed in $\text{OT}\sigma$. 
Conversely, if $\sigma$ is too large, the regularization interpolates toward MMD, removing the unique benefits of using $S_{\sigma}(\mu, \nu)$. 
To address this, we adopt a unique, dynamic regularization per epoch of training $\ell$, $\sigma_\ell$, where the transport plan is continually updated. 
We compute $\sigma_\ell$ iteratively as:
\begin{equation}
    \label{eqn:sigma_ell}
    \sigma_\ell = \max \left( 0.05 \cdot \max_{i,j} \| z_i - z^*_j \|_2, \; 0.01 \right) \;.
\end{equation}
In this formulation, $\sigma_\ell$ is dynamically adjusted based on the maximum pairwise distance $D_{ij} = ||z_i - z_j^*||_2$ between the source and target latent distributions.
We additionally set scaling based on the appropriate measures on the unit square or cube, which justifies the prefactor of $0.05$ in \myequation{}~\ref{eqn:sigma_ell} \citep{feydy2018interpolatingoptimaltransportmmd}.
This is further stabilized through the layer normalization of the latent vectors prior to computing $D_{ij}$.
This stabilization discourages outliers from disproportionately affecting the computation of $\sigma_\ell$. 
Finally, we impose a lower bound of $\sigma_\ell \geq 0.01$ to mitigate numerical instabilities and excessive computation as $S_\sigma$ approaches the unregularized Wasserstein distance.

Batches of size $n$ of latent vectors from source and target data, denoted $z_n$ and $z_n^*$, are retrieved through a forward pass of a single combined batch of the source and target data, $\mathbf{X} = [x_n, x_n^*]$, through the NN. 
The NN outputs a combined batch of latent vectors, denoted as $\mathbf{Z}$, which is subsequently separated into two subsets: one corresponding to the source domain and the other to the target domain.
This is particularly important for NNs, which utilize batch normalization \citep{ioffe2015batchnormalizationacceleratingdeep}. 
If $z_n$ and $z_n^*$ were passed separately, the batch statistics would be computed independently, leading to inconsistent normalization as the $z_n$ batch statistics will not incorporate $z_n^*$ and vice versa.

\begin{figure*}
    \label{fig:pipeline}
    \centering
    \includegraphics[width=\linewidth]{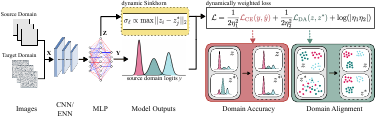}
    \caption{\algoname{} pipeline. The source and target domain batches of size $n$, $x_n$ and $x_n^*$, are first concatenated into a single batch $\mathbf{X}$ before being passed into the model. 
    After passing through the convolutional layers, the neural network produces a combined batch of latent vectors, $\mathbf{Z}$, extracted from the final linear layer. 
    This layer is positioned just before the output layer, which generates the class probabilities, $\mathbf{Y}$.
    Both $\mathbf{Z}$ and $\mathbf{Y}$ are split into separate batches for the source and target domains, resulting in $z_n$ (source) and $z_n^*$ (target) from $\mathbf{Z}$, and $y_n$ (source) and $y_n^*$ (target) from $\mathbf{Y}$, respectively.
    Only the source $y_n$ are used in training, as there are typically no target domain labels. 
    Both $z_n$ and $z_n^*$ are used to compute $\sigma_\ell$, a parameter that iteratively updates the regularization of the Sinkhorn plan in $\mathcal{L}_\text{DA}$.
    This process aligns the latent distributions of the source and target domains.
    This loss contribution is appropriately weighted with the classification loss, $\mathcal{L}_\text{CE}$, using a dynamic weighting of the tasks. 
    The result of training using \algoname{} is improved classification accuracy in both domains due to the aligned latent distributions, which can be visualized using non-linear clustering algorithms on the NN latent distributions.}
    \label{fig:enter-label}
\end{figure*}

%

We name this combined approach, which is dynamically adjusting a balance between cross-entropy and DA loss terms, and likewise dynamically adjusting the regularization of the Sinkhorn plan that facilitates the DA, \algoname{}. 
\algoname{} not only facilitates effective alignment between source and target domains, as we will demonstrate, but also reduces the need for extensive hyperparameter tuning --- a common challenge in DA implementations \citep{saito2021tunerightwayunsupervised}. 
As a result, this method provides a more automatic and reliable DA implementation with minimal computational overhead, leveraging existing resources that allow efficient computation of Sinkhorn divergences \citep{feydy2020fast}. 
We implement this technique using the \texttt{geomloss} library \citep{feydy2019interpolating}, which provides GPU implementations for Sinkhorn divergences and compatibility with \texttt{PyTorch} \citep{paszke2019pytorchimperativestylehighperformance}. 
A pipeline illustrating the dynamic DA approach during training is given in \myfigure{}~\ref{fig:pipeline}.

\subsection{The Jensen-Shannon Distance}
\label{sec:methods:jensen_shannon}

Recent advancements in DA theory have introduced the Jensen-Shannon (JS) divergence \citep{jsdiv} as a fundamental tool for understanding the inherent limitations of DA \citep{SHUI2022109808}. 
The JS divergence is a symmetrized statistical distance metric. 
For two latent distributions $\mu$ and $\nu$, the JS divergence \( D_{\text{JS}} \) is defined as
\begin{equation}
    D_{\text{JS}}(\mu \| \nu) = \frac{1}{2} D_{\text{KL}}(\mu \| \tau) + \frac{1}{2} D_{\text{KL}}(\nu \| \tau) \;,
\end{equation}
where \( D_{\text{KL}} \) denotes the KL divergence \citep{Kullback_1951}, defined as
\begin{equation}
    D_{\text{KL}}(\mu\;||\;\nu) = \int_{-\infty}^{\infty} p(z) \log \left( \frac{p(z)}{q(z)} \right) dz \;.
\end{equation}
Here, $p$ and $q$ denote probability densities of the latent features $z$ in the source and target distributions $\mu$ and $\nu$.
\( \tau = \frac{1}{2}(\mu + \nu) \) represents the mixture distribution of $\mu$ and $\nu$. 
The JS divergence offers two key advantages over \( D_{\text{KL}} \): it is symmetric (since, in general, \( D_{\text{KL}}(\mu \| \nu) \neq D_{\text{KL}}(\nu \| \mu) \)), and it is always finite. 
Additionally, the square root \( \sqrt{D_{\text{JS}}} \) defines a metric known as the Jensen-Shannon distance.

For DA applications, there exists a lower bound on the target domain loss~\citep{SHUI2022109808}: 
\begin{equation}
    \mathcal{L}_{t}(z^*) \geq \mathcal{L}_{s}(z) - \sqrt{D_{\text{JS}}(\mu \| \nu)} \;,
\end{equation}
where $\mathcal{L}_{s}$ is the source domain loss, $\mathcal{L}_{t}$ is the target domain loss, and $\sqrt{ D_{\text{JS}}(\mu \| \nu)}$ is the JS distance between the source and target latent distributions $\mu$ and $\nu$, respectively. 
This bound emphasizes that perfect alignment between source and target domains is fundamentally constrained by two components: \( \mathcal{L}_{s} \) and the JS distance between the source and target distributions. 
A smaller JS distance implies that the feature distributions of the source and target domains are closely aligned, which lowers the bound on \( \mathcal{L}_{t} \) and enables better transferability of the learned model. 
Conversely, a larger JS distance indicates a greater discrepancy, limiting the potential for minimizing target domain loss through adaptation alone.

The similarity between the source and the target latent distributions $\mu$ and $\nu$ is inherently influenced by the feature extraction capabilities of the neural network. 
DA methods aim to align features in the latent distribution; however, these features are ultimately limited by the network architecture. 
As a toy example, consider the case of image classification, where the architecture is a multi-layer perceptron (MLP). 
Many image classification tasks exhibit translation invariance, inherent in CNNs, but not in MLPs. 
DA on this task with CNNs will likely be more successful than with MLPs, as the translation invariance of the CNN further restricts the allowable features, and thus, the cost of alignment will be smaller.
In particular, we define ``robust'' features as those that respect the underlying data symmetries. 
More specifically, they yield similar classification probabilities under isometries that preserve the symmetries of the images.
If these symmetries persist in both the source and target domains, which is typically true except for extreme symmetry-breaking perturbations, then the cost of aligning robust features will be less than features learned from symmetry-agnostic architectures.
Consequently, it is reasonable to expect that ENNs endowed with appropriate higher-order symmetries will exhibit greater robustness and achieve more precise DA alignment than CNNs, because the latent distributions of ENNs are inherently more constrained.
Furthermore, when the assumption of underlying symmetries holds in both the source and target domains, as is typical in many DA applications, this advantage of ENNs becomes even more pronounced.

\section{Data}
\label{sec:data}

We evaluate the performance of our method on three simulated datasets and two real datasets: (1) a single-channel dataset of shapes consisting of lines, circles, and rectangles; (2) a single-channel dataset resembling astronomical objects, including stars, spirals, and elliptical galaxies; (3) the multichannel MNIST-M dataset \citep{ganin2016domainadversarialtrainingneuralnetworks}; (4) the Galaxy Zoo (GZ) Evo dataset of galaxies observed by two different optical telescopes~\citep{walmsley2024scalinglawsgalaxyimages}; and (5) the Multi-Modal Remote Sensing
Scene Classification (MRSSC2) dataset that contains optical and synthetic aperture radar (SAR) imaging~\citep{mrssc2}. 
The MRSSC2 dataset enables us to test performance in the presence of more severe covariate shifts caused by different wavelengths of the imaging data. 
The shapes and astronomical objects datasets are constructed using \texttt{DeepBench} \citep{osti_1989920}.
All datasets used in our experiments can be found on~\href{10.5281/zenodo.15215272}{Zenodo}. 

\subsection{Covariate Shifts}

\begin{figure*}
    \centering
    \includegraphics[width=\linewidth]{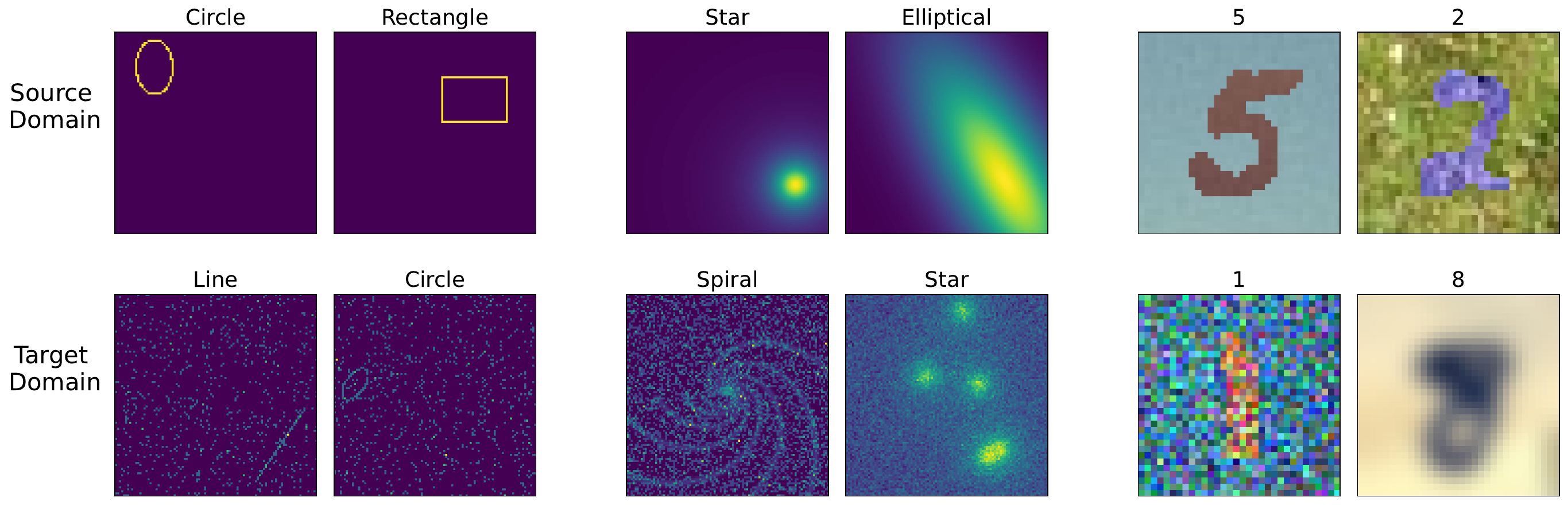}
    \caption{
    Example images for simulated datasets in the source domain (top row) and the target domain (bottom row) with corresponding labels. 
    \textbf{Left Panels:} Shapes dataset, featuring lines, rectangles, and circles, simulated with \texttt{DeepBench}. 
    This dataset includes variations in object positions and orientations, with Poisson noise added and normalized relative to the image signal in the target domain. 
    \textbf{Middle Panels:} Astronomical objects dataset, generated using \texttt{DeepBench}. 
    Parameters for spiral and elliptical galaxies were randomly sampled to determine morphology and position, while stars were generated similarly, with the number of stars as an additional parameter. Target domain images include additional Poisson noise.
    \textbf{Right Panels:} MNIST-M dataset with simulated Poisson noise (bottom left) and PSF blurring (bottom right) in the target domain.
    }
    \label{fig:dataset}
\end{figure*}

We use images from three simulated datasets, shown in~\myfigure{}~\ref{fig:dataset}, to study the performance on induced covariate shifts between the source and target domains. 
For all of our simulated datasets, we introduce fixed levels of Poisson noise in the target domain. 
Additionally, for MNIST-M, we also study the effects of PSF blurring in the target domain. 
By studying these two distinct covariate shifts, we evaluate the robustness of our method on covariate shifts relevant to data in realistic settings, particularly in the context of astrophysics and cosmology.

This is implemented for an image $I$ with grid values $(\zeta,\xi)$ and channels $c$ as
\begin{equation}
    I_{\text{Poisson}}(\zeta, \xi, c; S) = I(\zeta, \xi, c) + P\left(\frac{\langle I \rangle}{S} - \langle I \rangle\right) \;, 
\end{equation}
where $S$ is the signal-to-noise ratio, and $P$ denotes the Poisson distribution with rate parameter $\lambda = \frac{\langle I \rangle}{S} - \langle I \rangle$.
We incur PSF noise in each image channel by convolving the images with a Gaussian kernel \( G \) of kernel width \( \epsilon \):
\begin{equation}
    I_{\text{PSF}}(\zeta,\xi) = (I * G)(\zeta,\xi),
\end{equation}
where
\begin{equation}
    G(\zeta,\xi) = \frac{1}{2 \pi \epsilon^2} \exp \left( - \frac{\zeta^2 + \xi^2}{2 \epsilon^2} \right) \; .
\end{equation}

\subsection{Simulated Images}
\label{sec:data:simulated}

For the shapes dataset, we use \texttt{DeepBench} \citep{osti_1989920}, an open-source library for generating simulated datasets, and randomly construct rectangles, lines, and circles with varying radii (for circles), heights and widths (for rectangles), and lengths (for lines). 
The object positions, orientations, and thicknesses are also randomly assigned to introduce variance in the dataset. 
Poisson noise in the images is normalized with respect to the original image signal. 
We set a signal-to-noise ratio $S = 0.05$. 
Example images from the dataset can be seen in the left two panels of \myfigure{}~\ref{fig:dataset}.

We also use \texttt{DeepBench} for simulating astronomical objects. 
We generate astronomical objects resembling spiral galaxies, elliptical galaxies, and stars. 
For spiral galaxies, we randomly assign the centroid, winding number, and pitch to ensure morphological variation. 
The pitch is the angle indicating how tightly the arms are wound, while the winding number measures the total number of arm rotations from the center to the galaxy's edge. 
For elliptical galaxies, we vary the amplitude, radius, ellipticity, S\'ersic index \citep{sersic}, and rotation, as well as the centroid location. 
The amplitude sets the brightness level, ellipticity describes the degree of deviation from a circle, the S\'ersic index controls light concentration (higher values indicate more central concentration), and the rotation defines the orientation angle of the galaxy’s major axis. 
Lastly, we apply similar variations to generate stars, with the number of stars in each image uniformly distributed in the range $[0, 10]$. 
We use a fixed Poisson noise level of $S = 0.2$ to generate noisy target domain images.
This level of noise was chosen as it allows for a sufficient decrease in target domain performance for models without DA.
Example images are shown in the middle two panels of \myfigure{}~\ref{fig:dataset}.
Both of these datasets contain 12,000 training images (with $20\%$ being used for validation) and 3,000 test images in each domain. 
The images are square with 100 pixels on each side, and they are single-channel: each sample image has dimensions $100 \times 100 \times 1$.

MNIST-M \citep{ganin2016domainadversarialtrainingneuralnetworks} is a dataset that combines the handwritten digits of MNIST \citep{deng2012mnist} with randomly extracted color photos from BSDS500 \citep{amfm_pami2011} as background images. 
The original dataset contains 59,001 training images and 90,001 test images, out of which we use a balanced subset of 15,000 training (with $20\%$ set aside for validation) and 5,000 testing images in each domain. 
Since this is a three-channel dataset, images have a dimension of $32\times32\times3$.
We then create two types of target domain covariate shifts: 1) we set a signal-to-noise ratio of $S = 0.05$ for Poisson noise, and 2) a kernel width of $\epsilon =2$ for PSF blurring. Example images are shown in the right two panels of~\myfigure{}~\ref{fig:dataset}.

Both the images and the induced covariate shifts are simulated and do not capture all the complexities of real-world noise. 
Nevertheless, these datasets provide valuable benchmarks for challenges commonly encountered in astronomical and cosmological contexts, where DA methods can substantially enhance the robustness of neural network-based image classification pipelines.

\subsection{Real-Sky Galaxy Image Dataset}

\begin{figure*}
    \centering
    \includegraphics[width=\linewidth]{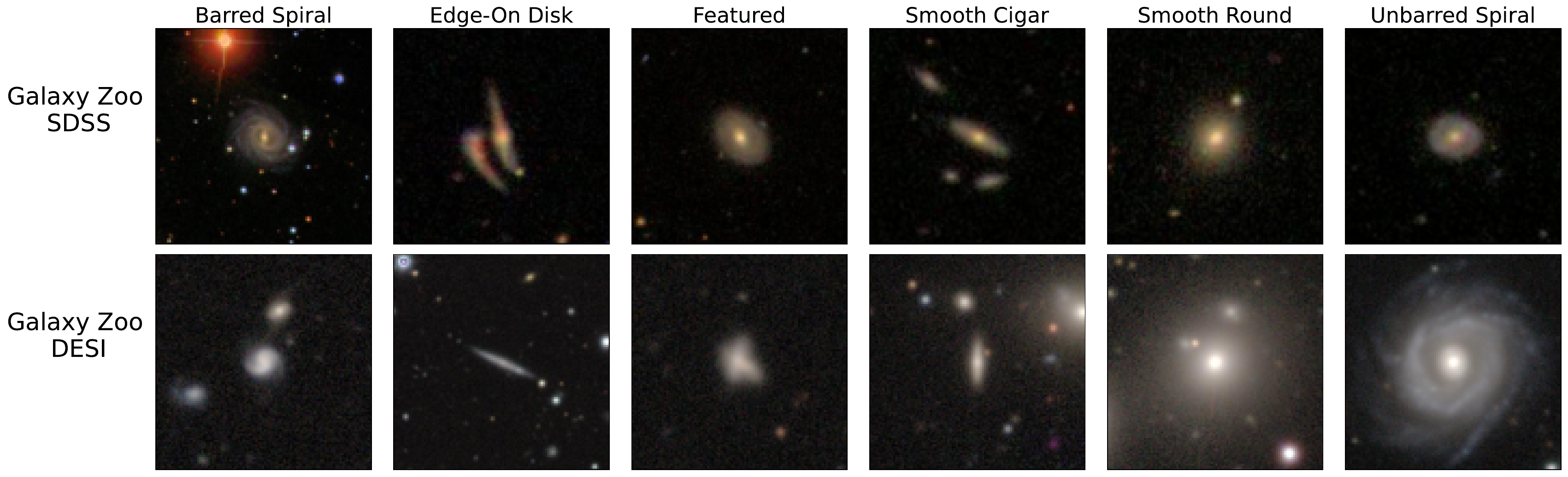}
    \caption{
    \textbf{Top Panel:} Example source domain images from the GZ Evo dataset with corresponding labels. 
    Images are from GZ2 data observed by SDSS.  
    \textbf{Bottom Panel:} Example target domain images from the GZ Evo dataset with the same labels. Images are from GZ DESI (combined observations from the DESI Imaging Surveys). 
    }
    \label{fig:galaxydataset}
\end{figure*}

We use the GZ Evo dataset \citep{walmsley2024scalinglawsgalaxyimages} to test cross-domain robustness in a more realistic scenario, where the covariate shift is present due to differences between images from two different astronomical surveys. These differences are due to different levels of observational noise, PSF blurring, pixel scale, as well as differences in populations of observable astronomical objects (how distant or how faint a resolved object can be).
GZ is a citizen science project that labels galaxy images through online participation.  GZ Evo combines labeled image datasets across several surveys and iterations of GZ. 
Within GZ Evo, we use the GZ2 Dataset from the Sloan Digital Sky Survey (SDSS) \citep{gzsdss} as the source, and a GZ Dark Energy Spectroscopic Instrument (DESI) dataset that combines observations from the DESI Imaging Surveys (DECals, MzLS, BASS, DES) \citep{gzdesi1, gzdesi2} as the target. 
Older GZ SDSS data contains objects up to magnitude $17$ in the $r$ band, and redshifts below $0.25$, while newer GZ DESI includes fainter objects up to magnitude $19$ and more distant objects with redshifts below $0.4$. Additionally, these surveys are different in the amount of observational noise and PSF blurring. 
Finally, GZ SDSS images include 3-filter $gri$ images, while GZ DESI includes 3-filter $grz$ images.

The original GZ Evo dataset contains $664,219$ images, each labeled according to vote counts (e.g., ``8 of 10 volunteers answered Spiral, and 7 of 10 answered Bar''). 
GZ Evo also offers an aggregated version where the vote counts are converted into distinct classes, which is convenient for developing machine learning models.
$239,408$ galaxies could be confidently assigned a distinct class based on the original vote counts.
Of this sample, $82,185$ corresponded to GZ DESI and $95,703$ to GZ SDSS.
The galaxy dataset used in this work contains a random sample of $40,000$ (32,000 training images with $20\%$ set aside for validation, and 8,000 testing images) images in each of the domains, across six distinct classes: ``smooth-round'', ``smooth-cigar'', ``unbarred spiral'', ``edge-on-disk'', ``barred spiral'', and ``featured'' galaxies. 
The ``featured'' galaxy class corresponds to any galaxy without a clear spiral structure or a visible bar, but which is also not completely smooth. 
The original galaxy images had dimensions $428 \times 428 \times 3$ and were subsequently downsampled to $100 \times 100 \times 3$ for more efficient training. 
Example images across all classes between GZ SDSS and GZ DESI are shown in \myfigure{}~\ref{fig:galaxydataset}.

\subsection{Remote Sensing
Scene Classification Dataset}

\begin{figure*}
    \centering
    \includegraphics[width=\linewidth]{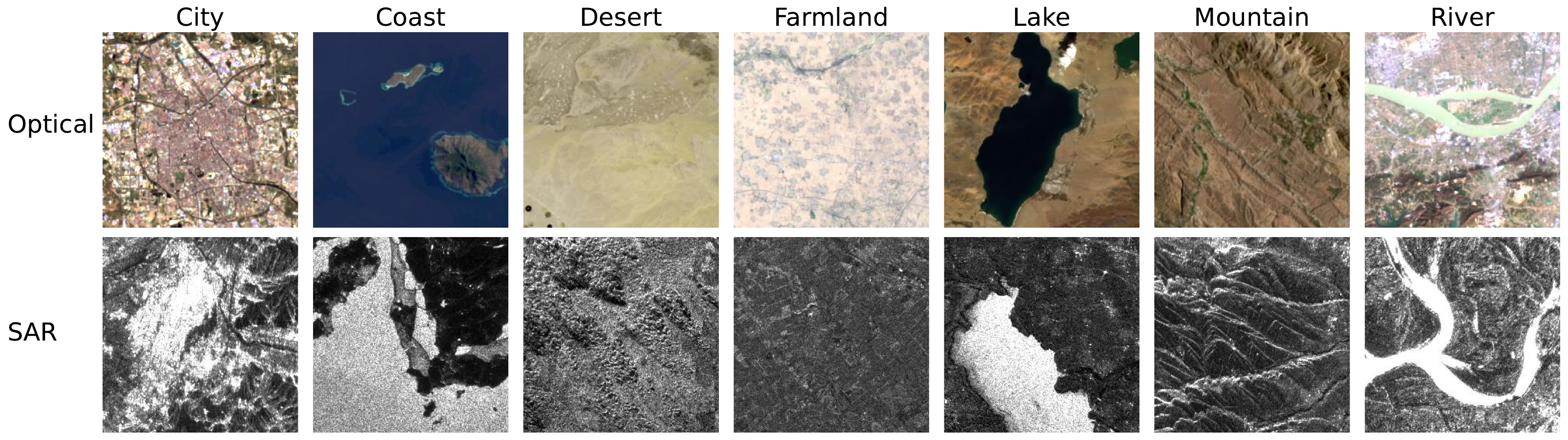}
    \caption{
    \textbf{Top Panel:} Example source domain optical images from the MRSSC2 dataset with corresponding labels.   
    \textbf{Bottom Panel:} Example target domain SAR images from the MRSSC2 dataset with the same labels. 
    }
    \label{fig:mrssc2dataset}
\end{figure*}

Finally, we test the efficacy of \algoname{} on a more extreme covariate shift problem, where the source and target domain images are coming from two different wavelength ranges.
Such applications are relevant in various scenarios, including medical imaging, remote sensing applications, and astronomy.

We use the MRSSC2 dataset, which contains multi-channel images across four wavelengths: optical, short wavelength infrared, thermal infrared, and SAR (microwave/radio wavelengths)~\citep{mrssc2}. 
Optical images show the true surface state, short wavelength infrared is more sensitive to soil moisture, thermal infrared reflects the surface temperature state, and SAR reflects the degree of surface backscattering. 
This can lead to substantial differences between images observed at different wavelengths. Together, the four imaging modalities (i.e., wavelengths) contain a total of $26710$ images split across seven scenes, including ``city,'' ``farmland,'' ``mountain,'' ``desert,'' ``coast,'' ``lake,'' and ``river.''

We used $4924$ optical images (source domain) and $4809$ SAR images (target domain) for training, which were split into $80\%$ training and $20\%$ validation. 
For testing, there are $1231$ source domain images and $1203$ target domain images. Example images across all classes between optical and SAR are shown in \myfigure{}~\ref{fig:mrssc2dataset}.
We chose these two datasets as the differences between images visually appeared the most extreme, which should lead to the most extreme covariate shift between them. Finally, the original images have dimensions $256 \times 256 \times3$; we downsampled them to $100 \times 100 \times 3$ for more efficient training.

\section{Network Architectures and Experiments}
\label{sec:experiments}

We evaluate our method on two sets of NNs: (1) CNNs constructed in \texttt{PyTorch} and (2)  ENNs constructed in \texttt{escnn}, a PyTorch-based library for easy construction of ENNs. 
Details of architectures and training are presented in~\ref{app:networks}.
Most of our experiments use an ENN equivariant to $D_4$, following the results in our previous work \citep{pandya2023e2equivariantneuralnetworks}, as this level of equivariance is beneficial but not overly computationally expensive to train.
We also study the performance of our method as a function of group order for the dihedral group. 
The code used in this work is available on our~\href{https://github.com/deepskies/SIDDA}{GitHub}.

\subsection{Equivariant Neural Networks}
\label{sec:networks:equivariant}

The efficacy of DA is limited by the feature extraction capabilities of the NN and its performance on the source domain. 
For image classification tasks, CNNs are useful due to their translation invariance and locality. 
There are, however, often additional symmetries inherent in the data, such as rotational and reflection invariance, that can be leveraged to enhance feature extraction and improve performance.

ENNs are a subclass of CNNs that can exploit higher-order symmetries besides the typical translation equivariance of CNNs \citep{cohen2016groupequivariantconvolutionalnetworks, cohen2016steerablecnns}. 
Of interest for 2D images are symmetries of the Euclidean group $E(2)$ --- in particular, the 2D special orthogonal group $\text{SO}(2)$ and the orthogonal group $\text{O}(2)$ and its associated subgroups. 
These (sub)groups allow ENNs to inherit symmetries of the circle and N-gon, respectively. 
As $\text{SO}(2)$ and $\text{O}(2)$ are continuous, they contain an infinite number of irreducible representations and have associated challenges when constructing architectures. 
For this reason, the discrete subgroup of $\text{O}(2)$, the dihedral group $D_N$, is used in this work, which is straightforward to construct using open-source software, such as \texttt{escnn} \citep{cesa2022a, e2cnn}.

For image data (particularly astronomical data), rotational symmetry (with or without reflections) is typically inherent. 
For instance, galaxy morphologies are often invariant under rotations because there is no preferred reference frame in the Universe.
Learning these symmetries can be induced in typical CNNs through data augmentation during training \citep{hernándezgarcía2020dataaugmentationinsteadexplicit, wang2022dataaugmentationvsequivariant}. 
However, the output feature map $f_\text{out}$ with grid values $(\zeta,\xi)$ from the convolution with kernel $K$ with grid values $(i,j)$,  
\begin{equation} 
    f_{\text{out}}(\zeta, \xi) = \sum_{i,j} K(i, j) \cdot f(\zeta + i, \xi + j), 
\end{equation} 
is inherently only translation-equivariant. 
In contrast, group convolution \citep{cohen2016groupequivariantconvolutionalnetworks, kondor2018generalizationequivarianceconvolutionneural} in ENNs can be equivariant to any arbitrary group $G$:
\begin{equation} 
    f_{\text{out}}(g) = \sum_{h \in G} K(h^{-1}g) \cdot f(h) 
\end{equation}
for $g, h \in G$: $g$ and $h$ correspond to the transformation for which the output feature map $f_\text{out}(g)$ is computed. 
For example, if $G$ is the group of 2D rotations $\text{SO}(2)$, $g$ and $h$ represent specific rotation angles.
ENNs construct specialized filters $K$ that are equivariant to the desired symmetries.
Therefore, they learn transformation-invariant features that preserve the underlying symmetries in the data throughout training. 

Each NN operation --- convolution, activation, pooling, and dropout --- must be equivariant \citep{dropout}. 
Our ENN architectures have three convolutional blocks, each containing a group convolution (\texttt{R2Conv}), batch normalization (\texttt{InnerBatchNorm}), ReLU activation, max pooling (\texttt{MaxPoolPointwise2D}), and dropout (\texttt{PointwiseDropout}). 
The global group equivariance for the network is defined before the first convolutional layer. 
Following the convolutional layers, a group pooling operation aggregates over all the symmetry channels, an essential step for constructing invariant representations in the latent distribution as discussed in \cite{hansen2024interpretingequivariantrepresentations}. 
Finally, the ENN includes two linear layers that downsample the learned representation to the designated number of output classes for classification.

The latent vector of the NN is extracted before the last linear layer, with no dropout in the linear layers. 
The latent vectors for all our networks undergo a layer normalization \citep{ba2016layernormalization}, which serves to stabilize the latent distributions by standardizing the feature distributions across each sample, ensuring consistent scaling and preventing the activations from drifting to extreme values. 
We found that the layer norm is useful when employing \algoname{} during training for more stable computations of $\sigma_\ell$ as given in \myequation{}~\ref{eqn:sigma_ell}.

The architectures used in our experiments are based on the $D_N$ group, which exhibits reflection symmetry. In the majority of our experiments, we set $N=4$, as it provides notable benefits without imposing significant computational overhead during training.
Our CNN has the same architecture, except the network components are not equivariant.

\begin{algorithm*}
\caption{\algoname{} optimization step}
\label{alg:DA}
\begin{algorithmic}[1]
\renewcommand{\COMMENT}[2][.5\linewidth]{%
  \leavevmode\hfill\makebox[#1][l]{//~#2}}
\REQUIRE $x$: Source domain inputs, $\hat{y}$: Source domain labels, $x^*$: Target domain inputs, $n$: batch size in each of the domains, $\eta_1$, $\eta_2$: Dynamic weighting parameters
\WHILE{not converged}
    \STATE $\mathbf{X} \gets [x_n,\, x_n^*]$ \COMMENT{Concatenate inputs}
    \STATE $\mathbf{Y},\, \mathbf{Z} \gets \text{model}(\mathbf{X})$ \COMMENT{Compute logits and latent features}
    \STATE $z_n,\, z_n^* \gets \mathbf{Z}$ \COMMENT{Split features into source and target}
    \STATE $y_n,\, y_n^* \gets \mathbf{Y}$
    \COMMENT{Split predictions into source and target}
    \STATE $D_{ij} \gets \|z_i - z_j^*\|_2$, $\forall \; i,j \in \{1, \dots, n\}$
    \COMMENT{Compute pairwise distances}
    \STATE $\sigma_\ell \gets \max(0.05 \times \max_{i,j} D_{ij},\ 0.01)$ \COMMENT{Compute dynamic blur parameter}
    \STATE $\mathcal{L}_{\text{DA}} \gets \text{Sinkhorn}(z_n,\, z_n^*,\, \sigma_\ell)$ \COMMENT{Compute Sinkhorn loss}
    \STATE $\mathcal{L}_{\text{CE}} \gets \text{CrossEntropy}(y_n,\, \hat{y}_n)$ \COMMENT{Compute classification loss}
    \STATE $\mathcal{L} \gets \dfrac{1}{2 \eta_1^2} \mathcal{L}_{\text{CE}} + \dfrac{1}{2 \eta_2^2} \mathcal{L}_{\text{DA}} + \log(|\eta_1 \eta_2|)$ \COMMENT{Compute total loss}
    \STATE $\text{loss.backward()}$ \COMMENT{Compute gradients}
    \STATE $\text{clip\_grad\_norm\_}(\text{model.parameters()},\ 10.0)$ \COMMENT{Gradient clipping}
    \STATE $\eta_1 \gets \max(\eta_1,\ 1\text{e-}3)$ \COMMENT{Clip $\eta_1$}
    \STATE $\eta_2 \gets \max(\eta_2,\ 0.25 \times \eta_1)$ \COMMENT{Clip $\eta_2$}
    \STATE $\text{optimizer.step()}$ \COMMENT{Update model parameters}
\ENDWHILE
\end{algorithmic}
\end{algorithm*}

\subsection{Training}
\label{sec:networks:training}

We train all networks typically ($\mathcal{L} = \mathcal{L}_{\text{CE}})$ and with \algoname{} (\myequation{}~\ref{eqn:DOT}) and study performance differences in the source and target domain for the two techniques. 
In all experiments, we use the AdamW optimizer \citep{loshchilov2019decoupledweightdecayregularization} with an initial learning rate of $10^{-2}$, a weight decay of $10^{-3}$, and a batch size of 128. 
A multiplicative learning rate decay of $0.1$ is applied twice sequentially during training to stabilize convergence. 

For all experiments, we use data augmentation comprising random rotations, flips, and affine translations. 
For the CNN, the augmentation instills approximate equivariance to encourage the model to learn rotation-invariant features. 
This is done to prevent predisposing the ENNs from performing better, but it also encourages faster convergence for the CNN. 
This later allows a more fruitful comparison between the latent distributions between ENNs, which have inherent rotation invariance to discrete rotations, and the CNNs, which have approximate invariance \citep{hansen2024interpretingequivariantrepresentations}.

For experiments with DA, an initial warm-up phase is implemented, during which only classification tasks are trained (using only $\mathcal{L}_{\text{CE}}$). 
A similar $0.1$ multiplicative learning rate decay as in the case of experiments without DA is also used. 
Model-saving criteria are based on the following: for experiments without DA, we use the best validation loss on the source domain for classification; for DA experiments, we use the sum of the validation classification loss on the source domain and validation DA loss.
The warm-up phase is intended to predispose the model to be at an ideal location in the loss landscape before starting DA. 
Empirically, we found this beneficial. 
The duration of the warm-up phase was tuned for each experiment, ensuring that it was long enough that the models were performant on the source domain but short enough that there was no overfitting. 
It was also found that ENNs required a shorter warm-up phase than CNNs. 
For example, for the shapes dataset, a warm-up of five and ten epochs were used for the $D_4$ and the CNN models, respectively. 
For the GZ Evo dataset, the warm-up phase was 20 and 30 epochs for the $D_4$ and the CNN models, respectively.
All models were trained for a maximum of 50 epochs for the shapes and astronomical objects datasets, whereas training was extended to a maximum of 100 epochs for the MNIST-M and GZ Evo datasets.
All training was done on one NVIDIA A100-80GB GPU, with the most complex experiment requiring about one hour to train.
Algorithm \ref{alg:DA} presents the forward pass for training with \algoname{}.

\subsection{Calibration}

Despite the impressive predictive capability of NNs in classification tasks across various fields, many real-world applications of NN-based classifiers also consider the confidence of each output class. 
Specifically, many NN-based classifiers can be uncalibrated, wherein the predicted class probabilities can frequently misrepresent the true class likelihood and lead to under- or overconfident predictions. 
In data-sensitive or safety-sensitive settings---such as medicine and biology---proper model calibration is essential for deploying NN-based classifiers~\citep{medcalibration}.
Similarly, in cosmology, simulation-based inference (SBI) pipelines that rely on trained classifiers must ensure proper calibration to guarantee that the inferred likelihood ratios or posterior probabilities are accurate and trustworthy~\citep{Cole_2022}.

Calibration techniques vary from regularization during training (either through architectural choices or additional loss terms) to post hoc methods that scale predicted probabilities (see \cite{wang2024calibrationdeeplearningsurvey} for a review). 
DA-based methods, however, have traditionally not been considered in the realm of regularization methods for model calibration. 
We will show that including DA with \algoname{} not only improves accuracy (see  Sections \ref{sec:results:simulated} and \ref{sec:results:galaxy}), but also calibration.

We evaluate model calibration using the Brier score and the Expected Calibration Error (ECE). 
The Brier score is the mean prediction error over all the classes:
\begin{equation}
    \text{Brier Score} = \frac{1}{C} \sum_{i=1}^C (y_i - \delta_{i \hat{y}})^2,
\end{equation}
where $y_i$ is the NN-predicted score for each class $i \in C$, where $C$ is the number of classes, $\hat{y}$ is the true class label, and  $\delta_{i \hat{y}}$ is the Kronecker delta.
Thus, a lower Brier Score is indicative of better calibration.

The ECE is the weighted average of the absolute difference between accuracy and confidence over $V$ equally spaced confidence bins. 
For each bin $B_v$, where $v \in \{1, \dots, V\}$, the accuracy and confidence are calculated based on the predictions within that bin. 
The ECE is defined as
\begin{equation}
    \text{ECE} = \sum_{v=1}^V \frac{|B_v|}{W} \left| \mathrm{acc}(B_v) - \mathrm{conf}(B_v) \right|,
\end{equation}
where $|B_v|$ is the number of samples in bin $v$, $W$ is the total number of samples, $\mathrm{acc}(B_v)$ is the average accuracy in bin $B_v$, and $\mathrm{conf}(B_v)$ is the confidence (average estimated probability) in bin $B_v$. 
The ECE is thus a measure of how much model confidence aligns with the true distribution of classes, and a lower ECE indicates better calibration.

\subsection{Neural Network Latent Distributions}

The distributions over the source and target latent encodings, $z$ and $z^*$, are the fundamental objects used in DA techniques.
Probing the latent distributions can give crucial insights into the success and failure points of DA.
The dimensionality of latent distributions is typically too large for visualization and analysis (256 in experiments used in this work).
Therefore, dimensionality reduction techniques are often employed before visualizing the latent distributions.
Techniques like t-SNE and UMAP \citep{tsne, umap} use local metrics, like pairwise distances or nearest-neighbor graphs to preserve the structure of the data at small scales while embedding it into a lower-dimensional space.
However, local metrics, and therefore these techniques, are limited because they primarily focus on preserving relationships within small neighborhoods of the data, often at the expense of capturing global structures or long-range dependencies that are critical for understanding the overall geometry or topology of the dataset.
In contrast, the isomap \citep{Tenenbaum2000} is a non-linear dimensionality reduction technique that estimates the global geometry of a latent vector manifold by using information of the nearest neighbors for each point in the latent space. 

In this work, we use isomaps to visualize latent distributions, and we use the mean Silhouette score to quantify the inter-class (between clusters) and intra-class (within a cluster) distances and evaluate the quality of the clustering.  
The silhouette score is
\begin{equation}
s = \frac{1}{Q} \sum_{i=1}^Q \frac{b(i) - a(i)}{\max(a(i), b(i))},
\end{equation}
where $Q$ is the number of points; $a(i)$ is the intra-cluster distance, which is the mean distance between the $i$-th data point and all other points within the same cluster; and $b(i)$ is the inter-cluster distance, the mean distance between the $i$-th data point and points in the nearest neighboring cluster. 
The Silhouette score is in the range $[-1, 1]$, where values close to one indicate well-clustered data, values near zero indicate overlapping clusters, and negative values indicate that the data point may be assigned to the wrong cluster.

\section{Results}
\label{sec:results}

All results are computed from three trained NNs, each with a different random seed for initializing weights. 
For each set of three trained networks, we estimate $1 \sigma$ uncertainties on our diagnostic metrics. 
We refer to a model trained without DA (i.e., only with cross-entropy loss) as ``<model>'' --- e.g., ``CNN'' or ``$D_4$''.
In contrast, we refer to a model trained with  \algoname{}  as ``<model>-DA'' --- e.g., ``CNN-DA'' or ``$D_4$-DA''.

\subsection{Simulated Datasets}
\label{sec:results:simulated}

\begin{table*}[!ht]
\centering
\caption{Classification accuracies for different model configurations on all datasets.}\label{tab:classification_accuracies}
\begin{tabular}{@{}ll|cc|cc@{}}
\toprule
\textbf{Dataset} & \textbf{Metric} & \textbf{CNN} & \textbf{CNN-DA} & \boldmath$D_4$ & \boldmath$D_4$\textbf{-DA} \\ \midrule

\multirow{2}{*}{Shapes} & Source Acc. (\%) & $99.80 \pm 0.04$ & $\mathbf{99.82 \pm 0.12}$ & $99.90 \pm 0.04$ & $\mathbf{99.92 \pm 0.02}$ \\
                        & Target Acc. (\%) & $50.47 \pm 8.39$ & $\mathbf{78.20 \pm 1.73}$ & $64.76 \pm 3.42$  & $\mathbf{99.71 \pm 0.06}$ \\ \midrule
                        
\multirow{2}{*}{\shortstack{Astro. \\ Objects}} & Source Acc. (\%) & $\mathbf{99.34 \pm 0.21}$ & $95.32 \pm 1.57$ & $\mathbf{99.98 \pm 0.02}$ & $99.89 \pm 0.50$ \\
                                      & Target Acc. (\%) & $50.81 \pm 2.89$ & $\mathbf{91.33 \pm 1.41}$ & $66.41 \pm 2.07$ & $\mathbf{97.19 \pm 0.51}$ \\ \midrule

\multirow{2}{*}{\shortstack{MNIST-M \\ (Noise)}} & Source Acc. (\%) & $\mathbf{95.64 \pm 0.12}$ & $95.31 \pm 0.09$ & $97.30 \pm 0.30$ & $\mathbf{97.45 \pm 0.02}$ \\
                               & Target Acc. (\%) & $68.32 \pm 2.72$ & $\mathbf{76.24 \pm 1.12}$ & $70.31 \pm 0.96$ & $\mathbf{87.55 \pm 0.16}$ \\ \midrule

\multirow{2}{*}{\shortstack{MNIST-M \\ (PSF)}} & Source Acc. (\%) & $95.64 \pm 0.12$ & $\mathbf{95.66 \pm 0.13}$  & $97.30 \pm 0.30$ & $\mathbf{97.95 \pm 0.10}$ \\
                             & Target Acc. (\%) & $75.00 \pm 1.44$ & $\mathbf{85.68 \pm 1.66}$  & $77.85 \pm 1.31$  & $\mathbf{93.00 \pm 1.14}$ \\  \midrule
                        
\multirow{2}{*}{\shortstack{Galaxy Zoo \\ Evo}} & Source Acc. (\%) & $81.49 \pm 0.32$ & $\mathbf{81.57 \pm 0.82}$ & $86.65 \pm 0.31$ & $\mathbf{87.58 \pm 0.06}$ \\
                            & Target Acc. (\%) & $70.65 \pm 2.26$ & $\mathbf{77.54 \pm 0.62}$& $79.48 \pm 1.52$ & $\mathbf{83.13 \pm 0.53}$ \\ \midrule
\multirow{2}{*}{MRSSC2} & Source Acc. (\%) & $\mathbf{76.14 \pm 1.63}$ & $71.27 \pm 0.85$ & $\mathbf{88.71 \pm 0.48}$ & $88.06 \pm 0.57$ \\
        & Target Acc. (\%) & $31.28 \pm 3.78$ & $\mathbf{36.80 \pm 1.58}$& $45.30 \pm 4.32$ & $\mathbf{48.10 \pm 1.56}$ \\
         \bottomrule               
\end{tabular}
\end{table*}

\begin{figure*}
    \centering
    \includegraphics[width=\textwidth]{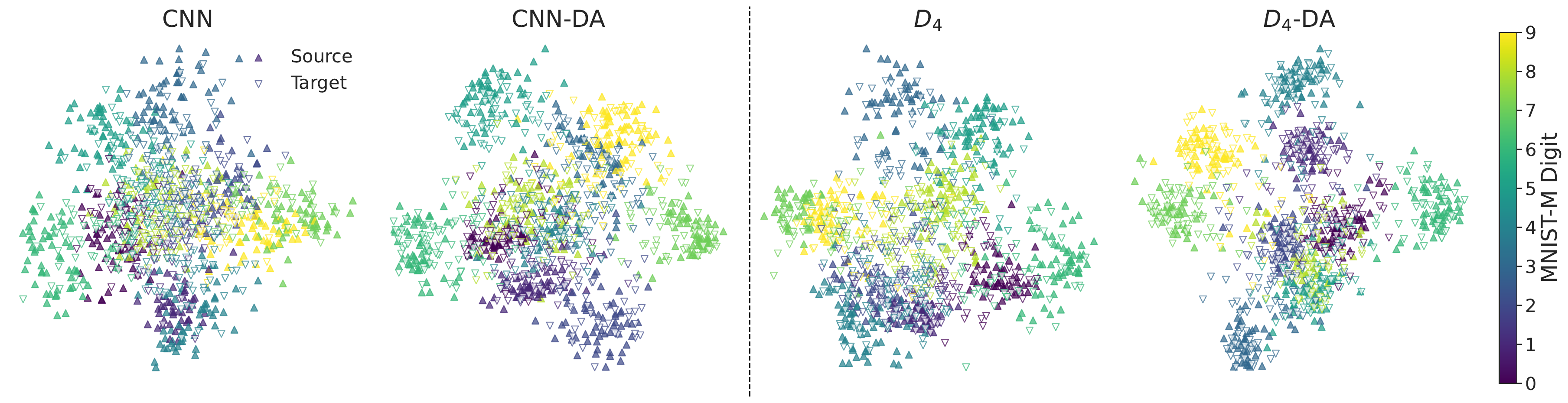}
    \caption{MNIST-M (Noise) latent distributions visualized with isomaps. Source (solid) and target (hollow) latent distributions are plotted atop each other to visualize latent distribution misalignment. 
    The inclusion of DA clearly improves the alignment of source and target latent distributions for both CNN and $D_4$ models. 
    It is also seen that the latent distribution of the $D_4$ is more clustered than the CNN, and even more so when $D_4$-DA and CNN-DA are compared. 
    The improved clustering and separation of classes in the latent space is suggestive of improved feature learning. }
    \label{fig:latent panel}
\end{figure*}

The test set accuracies for the CNN, $D_4$, CNN-DA, and $D_4$-DA models on the shapes, astronomical objects, MNIST-M, and GZ Evo datasets are shown in \mytable{}~ \ref{tab:classification_accuracies}. For the shapes and simulated astronomical objects dataset, both models achieve near-perfect accuracy (above $99\%$) on the source domain without DA, but classification accuracy is much lower on the target domain for both (between $50\%$ and $66\%$). With the inclusion of DA, the largest increase in target domain accuracy of $\approx40\%$ is for the astronomical objects datasets with CNN-DA.
Despite this, the $D_4$ model significantly outperforms the CNN in the target domain, in the case of both datasets. 
With DA, the $D_4$-DA model achieves greater than $97\%$ accuracy in both the source and target domains. 
While the CNN-DA shows a substantial improvement in target domain performance, the gap between the CNN-DA and $D_4$-DA remains considerable, with a difference of $21\%$ in target domain accuracy for the shapes dataset.

We test generalization capabilities in the presence of Poisson noise and PSF blurring on the MNIST-M dataset.
A similar trend emerges: the $D_4$ model is significantly more robust against both types of noise compared to the CNN. 
As in previous cases, with DA, neither model achieves the same performance as on the source domain (approximately $95\%$ for CNN-DA and $97\%$ for $D_4$-DA). However, the $D_4$-DA model demonstrates a greater potential for alignment than CNN-DA, achieving $93\%$ accuracy in the target domain during the PSF blurring experiments.

The results show that \algoname{} improves the source domain performance across most datasets for both CNN and $D_4$ models, except for the astronomical objects dataset for both models, and for the MNIST-M (noise) dataset for the CNN-DA model. In most cases, this drop in the source domain performance is below $1\%$ (except in the case of astronomical objects classified with the CNN-DA model), which is acceptable given that the inclusion of DA substantially improves the accuracy on the unlabeled target domain.
Additionally, the $D_4$-DA models converged to more similar performance across different seeds, as reflected by the small uncertainties in classification accuracy compared to CNN-DA.

We visualize the latent distributions of the CNN, $D_4$, CNN-DA, and $D_4$-DA in \myfigure{}~\ref{fig:latent panel} for the MNIST-M dataset with Poisson noise with the source (solid triangles) and target domain (hollow triangles) overlapped using isomaps \citep{Tenenbaum2000}. 
Without DA, there is significant overlap between different classes, particularly in the middle of the figure for both models, though it is more apparent for the CNN. 
This is reflected in the CNN and $D_4$ Silhouette scores (\mytable{}~\ref{tab:silhouette}), indicating similar levels of misclassified points, as seen in the target domain accuracy of approximately $68\%$ for CNN and $70\%$ for $D_4$.
With the inclusion of DA, the same classes from the source and target latent distributions for both models are better aligned. 
Furthermore, with DA, there is increased class separation, especially along the periphery, but there is still some overlap in the middle of the diagram (particularly for CNN-DA). 
Compared to the CNN latent distribution, the class separation for the $D_4$ latent distribution is much larger due to the equivariance in the $D_4$ model, which is reflected in the target domain Silhouette scores increasing more significantly for the $D_4$-DA when compared to the CNN-DA. 

\begin{table}[ht!]
\centering
\begin{tabular}{|l|c|c|}
\hline
\textbf{Model}           & \textbf{Source} & \textbf{Target} \\ \hline
CNN                & 0.1930          & -0.0539         \\ \hline
CNN-DA                & 0.3360          & 0.1150          \\ \hline
$D_4$                 & 0.2744          & -0.0247         \\ \hline
$D_4$-DA                    & \textbf{0.4023}          & \textbf{0.1983}          \\ \hline
\end{tabular}
\caption{Silhouette scores for CNN, $D_4$, CNN-DA, and $D_4$-DA on MNIST-M (Noise).}
\label{tab:silhouette}
\end{table}

\begin{figure}
    \centering
    \includegraphics[width=\linewidth]{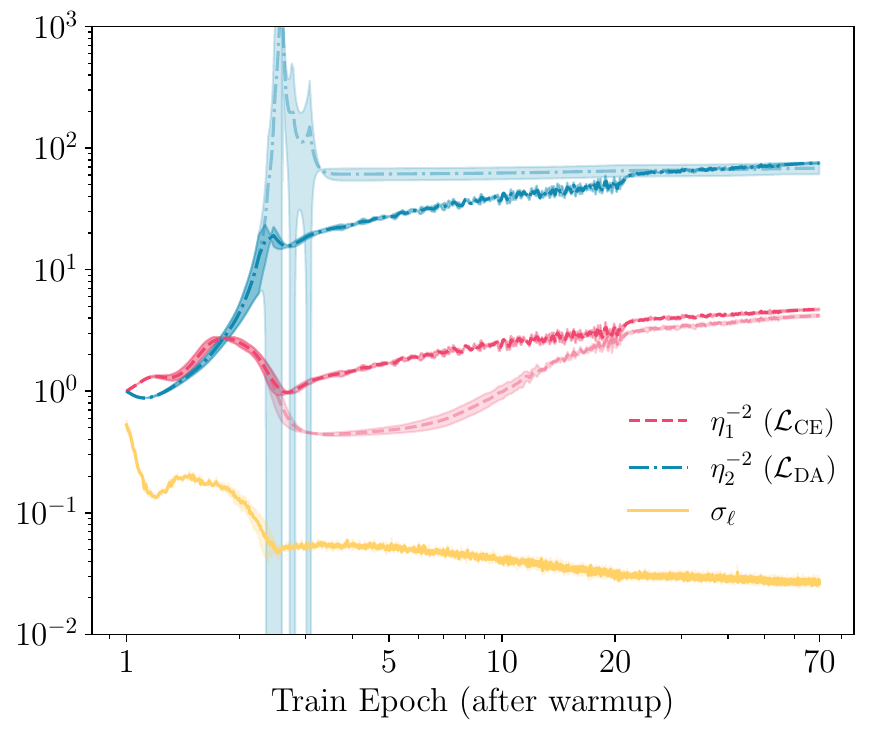}
    \caption{Evolution of the trainable coefficients, $\eta_1^{-2}$ and $\eta_2^{-2}$, and the Sinkhorn plan regularization strength $\sigma_\ell$ for CNN-DA trained on MNIST-M (Noise) after the $\mathcal{L}_\text{CE}$-only warm-up period. The shaded regions correspond to $1\sigma$ uncertainties from three training runs initialized with varying random seeds. With parameter clipping, as indicated in Algorithm \ref{alg:DA}, the time-evolution of both $\eta$'s is more stable, as indicated by the darker lines and their lower variance (i.e., narrower shaded regions).  Without parameter clipping (more transparent $\eta$ curves), the evolution of both $\eta$ parameters is more unstable with different random seeds, as indicated by the larger shaded regions and sharp increase of $\eta_2^{-2}$ around epoch three. Still, both $\eta$'s (with and without clipping) arrive at similar final values. It is also seen that the regularization strength of the Sinkhorn plan $\sigma_\ell$ continually decreases during training, as the NN latent spaces gradually become more aligned. This corresponds to the Sinkhorn plan more closely behaving as MMD at the beginning of training, while approaching the minimum allowed $S_{0.01}$ by the end.}
    
    \label{fig:paramevolution}
\end{figure}

We lastly study the evolution of the trainable loss coefficients $\eta_1$ and $\eta_2$, as well as the regularization strength of the Sinkhorn plan $\sigma_\ell$ during training. 
We show these parameters for the CNN-DA model trained on MNIST-M (Noise) in \myfigure{}~\ref{fig:paramevolution}; the overall behavior was typically seen in all model training. 
After the $\mathcal{L}_{\text{CE}}$-only warm-up phase, $\eta_2^{-2}$ (blue) quickly becomes greater than $\eta_1^{-2}$ (red), providing a larger weight for $\mathcal{L}_{\text{DA}}$. 
As training progresses, both $\eta_1^{-2}$ and $\eta_2^{-2}$ increase commensurately, indicating that a constant parameterization of loss coefficients throughout training is indeed not optimal, and that model convergence was aided by treating the coefficients as trainable parameters. The model convergence is also stable across seeds used during training, as indicated by the small shaded regions for $\eta_1$ and $\eta_2$ in the figure.
Upon looking at the loss curves, $\mathcal{L}_\text{DA}$ (without being weighted by $\eta_2^{-2}$) is roughly an order of magnitude smaller than $\mathcal{L}_\text{CE}$. 
After subsequent weighting of both terms with $1/ 2\eta_1^2$ and $1/2\eta_2^2$, the relative contribution of $\mathcal{L}_\text{DA}$ is still smaller than $\mathcal{L}_\text{CE}$; however, they are now roughly the same order of magnitude.
This allows the model to still prioritize the primary learning task (classification), while actively aligning the latent distributions with $\mathcal{L}_\text{DA}$. 
This was enforced by the clipping procedure for $\eta_2$ as shown in Algorithm \ref{alg:DA}, where the ratio must satisfy $\eta_2/\eta_1 \geq 0.25$.

We additionally study the evolution of $\eta_1^{-2}$ and $\eta_2^{-2}$ for the same experiment without enforcing the ratio for $\eta_2/\eta_1$. 
These are shown in the more transparent lines with the same corresponding colors in \myfigure{}~\ref{fig:paramevolution}.
We see that without clipping, $\eta_2^{-2}$ generally undergoes a sharp increase at the very beginning. 
Both $\eta$ values show a larger variance across seeds, with $\eta_2$ exhibiting significant fluctuations, as indicated by the more transparent blue shaded area in Figure \ref{fig:paramevolution}. 
By the end of training, the unclipped $\eta$ values both converge to similar values as their clipped counterparts. 
This indicates that the clipping enforced in \algoname{} stabilizes the evolution of these trainable loss coefficients.

We additionally see in \myfigure{}~\ref{fig:paramevolution} that $\sigma_\ell$ gradually decreases during training as the NN latent distributions continue to become more closely aligned. With \algoname{}, $S_\sigma$ initially interpolates more closely to MMD at the start of training, while by the end, it approaches the minimum allowed $S_{0.01}$, as shown in \myequation{}~\ref{eqn:sigma_ell}.

\subsection{Galaxy Zoo Evo Dataset}
\label{sec:results:galaxy}

We aim to address a common problem in real astronomical applications: the scenario where one has access to lower-quality, older observations with labels, but no labels exist for the newer, higher-quality target dataset from a more recent astronomical survey.
To test our model in this situation, we use galaxy datasets from GZ Evo~\citep{walmsley2024scalinglawsgalaxyimages}.
Namely, we use GZ2 from SDSS as the source domain and GZ DESI as the target domain.
This dataset is considerably larger than the previous simulated datasets. 
The results for these experiments are summarized in \mytable{}~\ref{tab:classification_accuracies}. 
Similar to the results for the simulated datasets, the $D_4$ model fully outperforms the CNN, with a $\approx9\%$ higher accuracy in the target domain. 
With DA, both CNN-DA and $D_4$-DA models have higher accuracy in both the source and target domains.
The performance difference between the two kinds of models is more moderate for the GZ Evo dataset than for the simulated datasets. 
This may be attributable to the GZ Evo data's larger size, greater morphological complexity, and the use of data augmentation during training.
In the large data limit, the inclusion of data augmentation causes CNNs to become approximately equivariant \citep{hernándezgarcía2020dataaugmentationinsteadexplicit, wang2022dataaugmentationvsequivariant}. 
Nevertheless, the $D_4$-DA achieves $\sim6\%$ higher accuracy in the target domain compared to the CNN-DA model.

The accuracy of all models in this experiment was lower than in the experiments with the simulated data, with no model achieving greater than $\approx88\%$ accuracy in either the source or target domain. 
This dataset is significantly larger than the others, so a deeper or larger model with additional training aids, such as residual connections, would likely yield better performance \citep{nakkiran2019deepdoubledescentbigger, he2015deepresiduallearningimage}.
Our goal is to study the efficacy of \algoname{}, and not to achieve state-of-the-art performance on this dataset, so we did not experiment with a more complex model.

\subsection{Robustness with Group Order}


\begin{table}[!ht]
\centering
\caption{Performance results for different orders of the dihedral group $D_N$, without and with DA.}\label{tab:grouporder}
\begin{tabular}{@{}lll@{}}
\toprule
\textbf{Group} & \textbf{Source Domain} & \textbf{Target Domain} \\ \midrule
$D_1$        & $96.03 \pm 0.18\%$ & $63.69 \pm 0.61\%$ \\
$D_2$        & $97.06 \pm 0.048\%$ & $67.88 \pm 2.5\%$ \\
$D_4$        & $97.30 \pm 0.28\%$ & $70.30 \pm 0.97\%$ \\
$D_8$       & $97.42 \pm 0.10\%$ & $71.67 \pm 0.28\%$ \\
$D_1$-DA   & $95.40 \pm 0.084\%$ & $75.35 \pm 0.71\%$ \\
$D_2$-DA   & $97.10 \pm 0.23\%$ & $84.98 \pm 1.9\%$ \\
$D_4$-DA  & $97.50 \pm 0.074\%$ & $87.70 \pm 0.26\%$ \\
$D_8$-DA  & $\mathbf{97.69 \pm 0.081\%}$ & $\mathbf{88.96 \pm 0.32\%}$ \\ \bottomrule
\end{tabular}
\end{table}

\begin{figure}
    \centering
    \includegraphics[width=\linewidth]{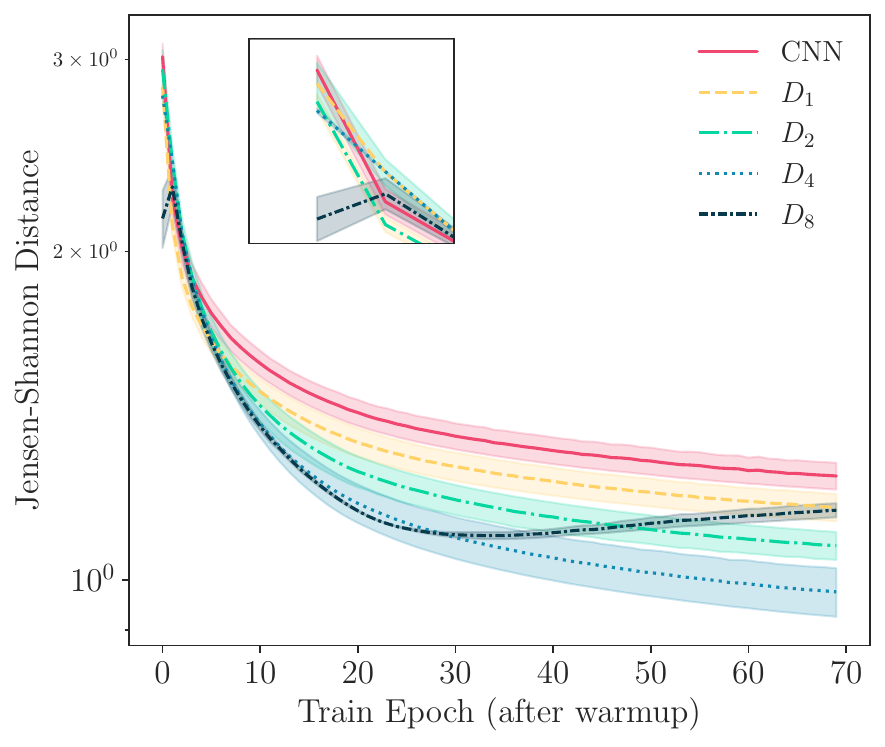}
    \caption{
    Jensen-Shannon (JS) distances for CNN-DA and $D_N$-DA ($N \in \{1,2,4,8\}$) models trained on MNIST-M (Noise). 
    Shaded regions correspond to $1\sigma$ uncertainties from three training runs initialized with varying random seeds.
    All models underwent a 30-epoch warm-up phase without DA, and the JS distance is shown for epochs thereafter, which is where \algoname{} is used. 
    Compared to the CNN, all $D_N$ models exhibit a lower JS distance between source and target domains after the warm-up phase, which can be attributed to the fact that the equivariance constraint encourages distributional similarity between the source and target latent distributions. 
    We see that the $D_N$ models also achieve more perfect alignment with the introduction of DA, as shown by the lower JS distances by the end of the training.
    This behavior correlates with the group order, except for $D_8$-DA, which achieved its best model much earlier in training and began overfitting. 
    }
    \label{fig:jsdistance}
\end{figure}

Next, we study the robustness of ENNs, both with and without \algoname{}, and with increasing orders of the dihedral group $D_N$, with $N \in \{1, 2, 4, 8\}$ on MNIST-M with Poisson noise in the target domain. 
The robustness of ENNs as a function of group order was previously studied in \citep{pandya2023e2equivariantneuralnetworks} for generalization to Poisson noise in images in the task of galaxy morphology classification. 
That work showed that robustness generally increased with group order, but high group-order models tended to overfit or have lower accuracy as a result of the equivariance constraint becoming too strong (see Figure 2 in \cite{pandya2023e2equivariantneuralnetworks}). 
Additionally, despite the robustness, there was no significant overlap in the latent distribution of the NNs when introduced to covariate shifts in the data. 
In the current work, we add to that previous study by examining the effect of \algoname{} on the robustness of ENNs across different group orders with the same data setup.

We follow a training procedure similar to that outlined in Section~\ref{sec:networks:training} for all networks. 
Models are still saved based on the best validation loss, and we show results from the best-performing models on the held-out test set. 
Source and target domain classification accuracies are shown in \mytable{}~\ref{tab:grouporder}. 
Both source and target accuracies increase with group order, with and without DA. 
There is also a slight increase in source domain accuracy when using DA, except for $D_1$, which contains a single reflection and therefore exhibits trivial equivariance.

We also track the evolution of the mean JS distance between source and target latent distributions for all models during training (after the initial warm-up phase has passed). 
Assuming perfect feature learning and optimal performance on the source domain, minimizing the JS distance between source and target domain latent distributions corresponds to optimal performance on the target domain \citep{SHUI2022109808}. 
As shown in \myfigure{}~\ref{fig:jsdistance}, after the initial 30-epoch warm-up phase, the JS distance decreases as the group order increases. 
Among the models, the $D_8$-DA model exhibits the lowest JS distance at the end of the warm-up, while the CNN-DA exhibits the highest.
This supports the claim that the constrained latent distribution of ENNs leads to significantly better alignment between source and target domains, as made in Section \ref{sec:methods:jensen_shannon}.

As training progresses with DA, the JS distance generally decreases with increasing group order, except for the $D_8$ model. 
This model begins to overfit 30 epochs after the warm-up phase concludes, as indicated by the JS distance in \myfigure{}~\ref{fig:jsdistance}. 
This was also confirmed upon inspection of the validation loss.
All other models reach their highest validation loss after epoch 90. 
Despite this overfitting, the best $D_8$-DA model still achieves the highest source and target domain accuracy, as shown in \mytable{}~\ref{tab:grouporder}. 

The overfitting observed in the $D_8$ model can be attributed to the stronger equivariance constraint (i.e., more weight sharing), which may limit the model's expressivity when the covariate shifts in the target domain do not fully respect the underlying symmetry. 
That is, the allowable space of features that respects the stronger equivariance will be inherently smaller than those respected by more lenient equivariance or the typical translation equivariance in CNNs, considering that ENNs assume perfect symmetries in the data. 
In the presence of perturbations, which is a typical case in the target domain for DA applications, this rarely holds. 
Solutions to relaxing the equivariance constraint while still enjoying the benefits of symmetry constraints have been extensively studied in other works \citep{wang2022approximatelyequivariantnetworksimperfectly, elsayed2020revisitingspatialinvariancelowrank}.

\subsection{Model Calibration}
\label{sec:calibration}

\begin{table*}[!ht]
\centering
\caption{Calibration metrics (expected calibration error and Brier score) for different model configurations.}\label{tab:calibration_metrics}
\begin{tabular}{@{}l|cc|cc@{}}
\toprule
\textbf{Metric} & \textbf{CNN} & \textbf{CNN-DA} & \boldmath$D_4$ & \boldmath$D_4$\textbf{-DA} \\ 
\midrule

\multicolumn{5}{@{}l}{\textbf{Shapes}} \\ 
\cmidrule(r){1-5}
Source ECE & $\mathbf{0.011 \pm 0.001}$ & $0.013 \pm 0.001$ & $0.011 \pm 0.002$ & $\mathbf{0.0074 \pm 0.0003}$ \\
Source Brier & $\mathbf{0.000734 \pm 0.000090}$ & $0.00112 \pm 0.00020$ & $0.000814 \pm 0.000200$ & $\mathbf{0.000349 \pm 0.000034}$ \\
Target ECE & $0.35 \pm 0.04$ & $\mathbf{0.29 \pm 0.004}$ & $0.20 \pm 0.03$ & $\mathbf{0.013 \pm 0.002}$ \\
Target Brier & $0.110 \pm 0.010$ & $\mathbf{0.0925 \pm 0.002}$ & $0.0564 \pm 0.009$ & $\mathbf{0.0015 \pm 0.0003}$ \\ 
\midrule

\multicolumn{5}{@{}l}{\textbf{Astronomical Objects}} \\
\cmidrule(r){1-5}
Source ECE & $\mathbf{0.041 \pm 0.010}$ & $0.075 \pm 0.020$ & $\mathbf{0.00695 \pm 0.00030}$ & $0.00899 \pm 0.00090$ \\
Source Brier & $\mathbf{0.00798 \pm 0.003}$ & $0.0220 \pm 0.006$ & $\mathbf{0.000132 \pm 0.000031}$ & $0.000746 \pm 0.000300$ \\
Target ECE & $0.17 \pm 0.04$ & $\mathbf{0.142 \pm 0.010}$ & $0.294 \pm 0.020$ & $\mathbf{0.053 \pm 0.008}$ \\ 
Target Brier & $0.0440 \pm 0.010$ & $\mathbf{0.0420 \pm 0.006}$ & $0.0804 \pm 0.009$ & $\mathbf{0.0150 \pm 0.003}$ \\ 
\midrule

\multicolumn{5}{@{}l}{\textbf{MNIST-M (Noise)}} \\
\cmidrule(r){1-5}
Source ECE & $0.161 \pm 0.003$ & $\mathbf{0.126 \pm 0.004}$ & $0.114 \pm 0.002$ & $\mathbf{0.0790 \pm 0.002}$ \\
Source Brier & $0.00991 \pm 0.00023$ & $\mathbf{0.00880 \pm 0.00030}$ & $0.00610 \pm 0.00030$ &  $\mathbf{0.00481 \pm 0.00010}$\\
Target ECE & $0.409 \pm 0.013$ & $\mathbf{0.355 \pm 0.009}$ & $0.390 \pm 0.020$ & $\mathbf{0.250 \pm 0.009}$ \\
Target Brier & $0.0450 \pm 0.003$ & $\mathbf{0.0370 \pm 0.002}$ & $0.0410 \pm 0.0008$ & $\mathbf{0.0210 \pm 0.0031}$ \\ 
\midrule

\multicolumn{5}{@{}l}{\textbf{MNIST-M (PSF)}} \\
\cmidrule(r){1-5}
Source ECE & $0.161 \pm 0.003$ & $\mathbf{0.124 \pm 0.004}$ & $0.114 \pm 0.002$ & $\mathbf{0.0750 \pm 0.002}$ \\
Source Brier & $0.00991 \pm 0.00023$ & $\mathbf{0.00850 \pm 0.00040}$ & $0.00610 \pm 0.00030$ & $\mathbf{0.00400 \pm 0.00020}$ \\ 
Target ECE & $0.384 \pm 0.013$ & $\mathbf{0.272 \pm 0.012}$ & $0.340 \pm 0.020$ & $\mathbf{0.181 \pm 0.001}$ \\ 
Target Brier & $0.0340 \pm 0.001$ & $\mathbf{0.0230 \pm 0.001}$ & $0.0270 \pm 0.003$ & $\mathbf{0.0130 \pm 0.00007}$ \\  
\midrule

\multicolumn{5}{@{}l}{\textbf{Galaxy Zoo Evo}} \\
\cmidrule(r){1-5}
Source ECE & $0.283 \pm 0.0019$ & $\mathbf{0.264 \pm 0.0044}$ & $0.2322 \pm 0.00086$ & $\mathbf{0.206 \pm 0.0011}$ \\
Source Brier & $0.0453 \pm 0.00031$ & $\mathbf{0.0439 \pm 0.0010}$ & $0.0341 \pm 0.00059$ & $\mathbf{0.0319 \pm 0.00014}$ \\ 
Target ECE & $0.324 \pm 0.0078$ & $\mathbf{0.301 \pm 0.0018}$ & $0.271 \pm 0.0042$ & $\mathbf{0.241 \pm 0.0026}$ \\ 
Target Brier & $0.0538 \pm 0.0015$ & $\mathbf{0.051 \pm 0.00029}$ & $0.0411 \pm 0.0012$ & $\mathbf{0.0382 \pm 0.00048}$ \\ 
\midrule
\multicolumn{5}{@{}l}{\textbf{MRSSC2}} \\
\cmidrule(r){1-5}
Source ECE & $\mathbf{0.331 \pm 0.00600}$ & $0.407 \pm 0.0121$ & $\mathbf{0.250 \pm 0.0152}$ & $0.304 \pm 0.0159$ \\
Source Brier & $\mathbf{0.0409 \pm 0.00113}$ & $0.0615 \pm 0.00247$ & $\mathbf{0.0243 \pm 0.00178}$ & $0.0328 \pm 0.00312$ \\ 
Target ECE & $\mathbf{0.422 \pm 0.00760}$ & $0.446 \pm 0.00418$ & $\mathbf{0.407 \pm 0.0212}$ & $0.452 \pm 0.0453$ \\ 
Target Brier & $\mathbf{0.0584 \pm 0.00248}$ & $0.0975 \pm 0.0000711$ & $\mathbf{0.0505 \pm 0.00510}$ & $0.0656 \pm 0.0141$ \\ 

\bottomrule
\end{tabular}
\end{table*}


For all datasets, we observe that the $D_4$ and $D_4$-DA models result in lower ECE and Brier scores across most experiments in both the source and target domains when compared to the CNN and CNN-DA. Across all experiments, the largest improvement is observed for the $D_4$-DA model applied to the shapes target domain data, where both the ECE and Brier score are reduced by more than an order of magnitude compared to the regular $D_4$ model. Furthermore, we see that in most experiments, the classification accuracies as given in \mytable{}~\ref{tab:classification_accuracies} are correlated with the calibration metrics in \mytable{}~\ref{tab:calibration_metrics} in both the source and target domains. That is, as classification accuracy in either the source or target domain increases, model calibration improves.

With the inclusion of DA, there is an improvement in the calibration scores for both data domains and for all models for the MNIST-M datasets, as well as GZ Evo data. 
Between these datasets, the largest improvement is for the $D_4$-DA model on the MNIST-M (PSF) dataset in the target domain, exhibiting an approximate factor of two reduction for both the ECE and Brier score.
However, at the same time, we observe a decrease in the source domain calibration for all DA experiments in the shapes and astronomical objects datasets, except for $D_4$-DA in the shapes dataset. 
This indicates that in the source domain, at least for simpler datasets, the inclusion of DA can potentially worsen the calibrations of the models. 
This decrease in performance is not always apparent in the (uncalibrated) accuracies, as shown in \mytable{}{}~\ref{tab:classification_accuracies}. 
For instance, the source accuracies with and without DA are very similar (within the margin of error) in all cases, except for the CNN-DA model applied to the astronomical objects dataset, where the source domain accuracy drops around $4\%$ with the inclusion of DA.
Nonetheless, in the target domain, we observe strict improvements in both accuracy and calibration across all experiments with the inclusion of DA with \algoname{}, as shown in \mytable{}~\ref{tab:classification_accuracies} and \mytable{}~\ref{tab:calibration_metrics}.

\subsection{Method Comparisons}
\label{sec:comparisons}

Next, we use the MNIST-M (Noise) dataset to compare \algoname{} to MMD and the Wasserstein distance in terms of accuracy, calibration, and computational efficiency (see \mytable{}~\ref{tab:mmd_wass_comparison}).
The latter two methods are distance-based and represent limiting cases of the Sinkhorn divergence. 
We use a Gaussian MMD with a fixed kernel width of $\epsilon = 0.05$ and approximate the true Wasserstein distance as the Sinkhorn divergence in the limit that $\sigma \rightarrow 0$, with a fixed regularization of $\sigma = 10^{-12}$.
All models were trained for 100 epochs using identical training hyperparameters (see \mytable{}~\ref{tab:training_configs}), model saving criteria, and hardware (NVIDIA A100 GPU).

\algoname{} outperforms both MMD and the Wasserstein distance in terms of target domain accuracy, ECE, and the Brier score, achieving the best overall performance. Specifically, it yields an approximate $1\%$ improvement over the Wasserstein distance and about $15\%$ over MMD in target domain accuracy.
The Wasserstein $D_4$ model achieves a $97.56\%$ accuracy on the source domain, almost within error bars of the \algoname{}-$D_4$ model, which has a $97.45\%$ accuracy.

{\algoname{} achieves the best calibration among all methods, including on the source domain, despite the Wasserstein distance performing slightly better in that setting. Overall, models based on the Wasserstein distance demonstrate the next best calibration, followed by those using MMD, as evidenced by the ECE and Brier scores. 
In all cases, the $D_4$ models achieve better calibration than the CNN models.

The wall clock time of \algoname{} is similar to that of MMD; both models require $<10$ minutes to train. 
The Wasserstein distance models require much more time --- e.g., 24 minutes for the Wasserstein-$D_4$ model. 
This can be attributed to the slower convergence for Sinkhorn iterations as the entropic regularization approaches zero ($\sigma \rightarrow 0$).

\begin{table*}[!ht]
\centering
\caption{Performance comparison of SIDDA with Gaussian MMD and Wasserstein distance DA methods on MNIST-M (Noise).} 
\label{tab:mmd_wass_comparison}
\resizebox{\textwidth}{!}{%
\begin{tabular}{@{}l|l|cc|cc|cc|c@{}}
\toprule
\textbf{Method} & \textbf{Model} & 
\multicolumn{2}{c|}{\textbf{Accuracy (\%)}} & 
\multicolumn{2}{c|}{\textbf{Expected Calibration Error (ECE)}} & 
\multicolumn{2}{c|}{\textbf{Brier Score}} & 
\textbf{\shortstack{Time \\ (min)}} \\
\cmidrule(r){3-4} \cmidrule(r){5-6} \cmidrule(r){7-8}
 &  & \textbf{Source} & \textbf{Target} & \textbf{Source} & \textbf{Target} & \textbf{Source} & \textbf{Target} &  \\ \midrule

\multirow{2}{*}{SIDDA}      
    & CNN           & $95.31 \pm 0.09$ & $76.24 \pm 1.12$ & $0.126 \pm .004 $ & $0.355 \pm .009 $ & $0.00880 \pm .00030$ & $0.0370 \pm 0.002$ & 8 \\
    & \boldmath$D_4$ & $97.45 \pm 0.02$ & $\mathbf{87.55 \pm 0.16}$ & $\mathbf{0.0790 \pm .002}$ & $\mathbf{0.250 \pm 0.009}$ & $\mathbf{0.00481 \pm 0.00010}$ & $\mathbf{0.0210 \pm 0.0031}$ & 10 \\ \midrule

\multirow{2}{*}{MMD}        
    & CNN           & $95.60 \pm 0.18$ & $68.90 \pm 1.52$ & $0.164 \pm 0.00686$ &  $0.409 \pm 0.00778$ & $0.0102 \pm 0.000523$ & $0.0459 \pm 0.00172$ & 5 \\
    & \boldmath$D_4$ & $97.25 \pm 0.04$ & $72.94 \pm 1.97$ & $0.122 \pm 0.00134$ & $0.382 \pm 0.000297$ & $0.00656 \pm 0.000112$ & $0.0390 \pm 0.000462$ & 7 \\ \midrule

\multirow{2}{*}{Wasserstein} 
    & CNN           & $94.83 \pm 0.19$  & $75.54 \pm 1.13$ & $0.143 \pm 0.00466$ & $0.371 \pm 0.00955$ & $0.0100 \pm 0.000456$ & $0.0389 \pm 0.00152$ & 19 \\
    & \boldmath$D_4$ & $\mathbf{97.56 \pm 0.05}$  & $86.62 \pm 0.24$ & $0.0938 \pm 0.00242$ & $0.273 \pm 0.00693$ & $0.00552 \pm 0.000141$ & $0.0232 \pm 0.000752$ & 24 \\ 
\bottomrule
\end{tabular}
}
\end{table*}

\subsection{Comparison with Fixed Loss Coefficients}

\begin{table}[!ht]
\centering
\caption{Source and target domain accuracies for different loss formulations for the CNN-DA model on MNIST-M (Noise).}
\label{tab:cnn_loss_ablation}
\resizebox{.5\textwidth}{!}{%
\begin{tabular}{@{}l|cc@{}}
\toprule
\textbf{Loss Formulation} & Source Acc. (\%) & Target Acc. (\%) \\ \midrule

$\mathcal{L}_\text{train}$   & $95.31 \pm 0.09$ & $\mathbf{76.24 \pm 1.12}$ \\
$\mathcal{L}_\text{C,D}$  & $95.35 \pm 0.13$ & $72.35 \pm 1.08$ \\
$\mathcal{L}_\text{C,10D}$  & $94.96 \pm 0.18$ & $74.88 \pm 1.00 $ \\
$\mathcal{L}_\text{10C,D}$  & $\mathbf{95.61 \pm 0.28}$ & $71.69 \pm 0.92$ \\

\bottomrule
\end{tabular}
}
\end{table}

Next, we compare \algoname{} with trainable loss coefficients $\eta_i$ (which in this experiment we will denote as $\mathcal{L}_\text{train}$) with fixed loss coefficients by examining three losses, each associated with a different fixed coefficient. 
Respectively, these put equal emphasis on each loss, favor the DA loss, and favor the classification loss: 
\begin{align}
    \mathcal{L}_\text{C,D} &= \mathcal{L}_\text{CE} + \mathcal{L}_\text{DA} \label{eqn:fixed}\\
    \mathcal{L}_\text{C,10D} &= \mathcal{L}_\text{CE} + 10 \cdot \mathcal{L}_\text{DA} \label{eqn:10da} \\
    \mathcal{L}_\text{10C,D}  &= 10\cdot\mathcal{L}_\text{CE} + \mathcal{L}_\text{DA}\label{eqn:10ce}.
\end{align}
For this set of experiments, we train the CNN models on MNIST-M (Noise). 
All models were trained with hyperparameters as shown in \mytable{}~\ref{tab:training_configs}, were saved according to the best validation loss, and the best model results are communicated in \mytable{}~\ref{tab:cnn_loss_ablation}.

\mytable{}~\ref{tab:cnn_loss_ablation} compares \algoname{} ($\mathcal{L}_\text{train}$) with the previously described loss formulations.
The highest source domain accuracy is achieved with $\mathcal{L}_\text{10C,D}$, whereas the highest target domain accuracy corresponds to $\mathcal{L}_\text{train}$. 
Still, $\mathcal{L}_\text{train}$ source domain accuracy is in $1\sigma$ agreement with that achieved with $\mathcal{L}_\text{10C,D}$ and exhibits a smaller error bar, suggesting more stable training. 
$\mathcal{L}_\text{train}$ also achieves better performance in the target domain when compared with $\mathcal{L}_\text{C,D}$, and with the DA-favored loss ($\mathcal{L}_\text{C,10D}$).
As evidenced by the high performance in both source and target domains, \algoname{} ($\mathcal{L}_\text{train}$) overall performs better than any of the models with fixed loss coefficients.
All models exhibit an approximate $1\%$ uncertainty in target domain accuracy, while \algoname{} has a slightly larger uncertainty in the target domain; this may be due to the trainable nature of the coefficients. 
Additional model initializations and experiments are necessary to draw more definitive conclusions regarding the stability of dynamic loss weighting. 

The optimal choice of fixed loss coefficients will change for each dataset, problem, and type of loss.
A poor choice of fixed loss coefficients can lead to models that perform poorly in the target domain when the DA loss is too small, or completely fail to learn the main objective when the DA loss is too large and prevents minimization of the main task loss. Additionally, suboptimal performance in both domains can occur if the DA loss is introduced at the wrong time during training.
Our findings demonstrate that the approach introduced by \citet{kendall2018multitasklearningusinguncertainty} allows one to avoid manual hyperparameter tuning for loss coefficients in DA tasks, while still achieving strong predictive performance on both source and target domains. 
This is particularly important for problems with very large datasets or complex models, where longer training times may be required.

\subsection{Application to Severe Covariate Shifts}
\label{sec:cross-mod}

We have thus far demonstrated the efficacy of \algoname{} across a range of datasets and covariate shifts---arising from factors such as varying noise levels, blurring, or differences in image quality (e.g., images captured by different telescopes within the same wavelength range).
More substantial covariate shifts can arise when the source and target datasets include images captured at different wavelengths.
We study the efficacy of \algoname{} on this type of covariate shift using the MRSSC2 dataset. 
We use optical observations as the source domain and SAR (which operates in cm wavelengths, i.e., microwave/radio range) images as the target domain.

As shown in ~\mytable{}~\ref{tab:classification_accuracies}, the efficacy of \algoname{} on the MRSSC2 dataset is considerably lower than the other datasets studied.
In both the CNN-DA and $D_4$-DA models, the target domain performance increases (up to $\sim5\%$).
However, in both cases, the target domain performance is considerably worse than the source domain, even with the inclusion of \algoname{}. 
Furthermore, even this small increase in target domain performance with DA causes a decrease in source domain accuracy. 
The source domain performance of the CNN-DA model is less than the CNN model; similarly, the $D_4$-DA model performance is lower than the $D_4$ model, but the accuracy drop is much lower (less than $1\%$).

\begin{figure*}
    \centering
    \includegraphics[width=1\textwidth]{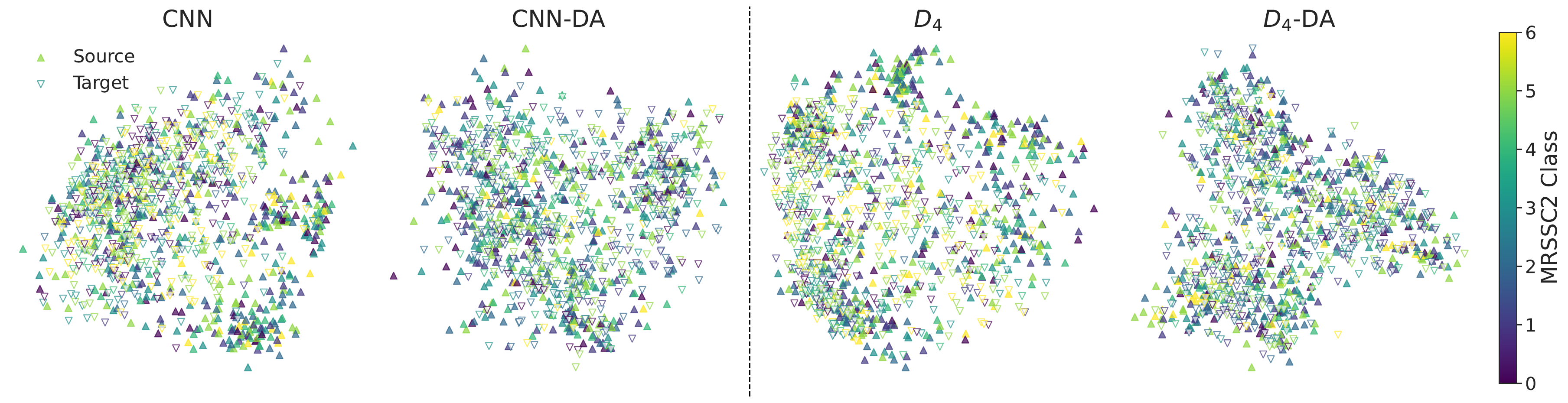}
    \caption{MRSSC2 latent distributions, visualized using isomaps, with the source domain shown as solid markers and the target domain as hollow markers. 
    Both the CNN and $D_4$ models exhibit substantially less clustering in the latent space---compared to the MNIST-M (Noise) dataset shown in~\myfigure{}~\ref{fig:latent panel}---which may account for the modest performance gains achieved by DA on this dataset.}
    \label{fig:mrssc2latent}
\end{figure*}

We use isomaps to visualize the latent distributions of the CNN, $D_4$, CNN-DA, and $D_4$-DA models in ~\myfigure{}~\ref{fig:mrssc2latent} for the MRSSC2 dataset, with the source (solid triangles) and target domain (hollow triangles).
Without DA, there is no well-defined clustering in any of the latent distributions, which can be indicative of poor feature learning. 
There is also significant misalignment between the source and target latent distributions. 
With DA, the misalignment decreases, but the latent space representations of different classes are not as defined as previously seen with well-performing \algoname{} models (see MNIST-M (Noise) latent distributions in \myfigure{}~\ref{fig:latent panel}). 
The poor feature learning even without DA may explain the modest performance improvement with DA on the MRSSC2 dataset, as shown in ~\mytable{}~\ref{tab:classification_accuracies}.

In addition, the inclusion of \algoname{} does not improve model calibration for both models on the MRSSC2 dataset, as seen in \mytable{}~\ref{tab:calibration_metrics}. 
Even in the target domain, where \algoname{} somewhat increased classification accuracy, the calibration decreases. 
This is in contrast with what was observed with our other datasets, where the domain alignment was successful, and the ECE and Brier scores generally improved for the CNN-DA and $D_4$-DA models.

The sub-optimal performance on the MRSSC2 dataset might be the result of larger differences between classes when observed at different wavelengths.  In this case, features extracted from the source domain images may be much less applicable to the target domain, which causes a considerably larger difference between source and target domain accuracy for the CNN and $D_4$ models (trained without DA) compared to other datasets studied.
These feature learning differences can then make domain alignment of latent representations near the end of the architecture --- as done here --- unfeasible.

For distance-based DA approaches like \algoname{}, this sub-optimality can potentially be addressed by intermediate domain alignment, similar to Deep Adaptation Networks (DANs)~\citep{long2015learningtransferablefeaturesdeep} and Joint Adaptation Networks (JANs)~\citep{long2017deeptransferlearningjoint}, and as indicated by results presented in \citet{mrssc2}. 
This would then impose domain alignment at all stages of the architecture, from the early convolutional layers, which perform general feature learning, to the later convolutional and linear layers, which learn more detailed and task-specific features. 
Additionally, one could approach more extreme domain shift problems with adversarial methods such as Domain Adversarial Neural Networks (DANNs)~\citep{ganin2016domainadversarialtrainingneuralnetworks}, which are more flexible and avoid explicit calculation of any distance metric. Finally, using more complex networks could enable better feature learning, leading to better clustering of different classes compared to what we see in Figure~\ref{fig:mrssc2latent}. This would enable easier domain alignment, potentially improving the efficacy of \algoname{} when applied to the MRSSC2 dataset.

\section{Conclusions and Outlook}
\label{sec:conclusion}

In this work, we introduced \algoname{} --- a semi-supervised DA approach that leverages principled methods for optimal alignment of NN latent spaces and training with multiple loss terms.
This is in contrast to most DA applications, which face the challenge of requiring extensive hyperparameter tuning, making training NNs time-consuming, expensive, or unfeasible. 
\algoname{} is an ``out-of-the-box'' DA method that is applicable in various domains, and it requires labeled training data only in the source domain, not in the target domain. 

Our method relies on the Sinkhorn divergence, a symmetrized variant of regularized OT distances that corrects for the bias in $\text{OT}_\sigma$. 
In particular, our method dynamically adjusts the strength of regularization on a per-epoch basis, dependent on the pairwise distance between the source and target latent distribution vectors. 
In addition, we use dynamic weighting of the cross-entropy and DA loss terms on a per-epoch basis, which ensures a balance in training and improves predictive performance in both the source and target domains.

We test our method on shapes and astronomical objects datasets simulated using \texttt{DeepBench}, the MNIST-M dataset, a dataset of real galaxy images from Galaxy Zoo, and the MRSSC2 remote sensing dataset. 
These experiments encompass covariate shifts induced by Poisson noise, PSF blurring, differences between real optical astronomical surveys, and differences between images observed at different wavelengths.

We draw the following conclusions about \algoname{}:

\begin{itemize}
    \item \algoname{} requires minimal hyperparameter tuning to achieve a considerable increase in target domain performance, and can perform better than traditional fixed-hyperparameter use cases, as shown in ~\mytable{}~\ref{tab:cnn_loss_ablation};
    \item \algoname{} combines the performance of the Wasserstein distance with the efficiency of MMD, shown through comparisons presented in \mytable{}~\ref{tab:mmd_wass_comparison};
    \item \algoname{} is compatible with various models, including CNNs and ENNs, which are equivariant to different groups. 
    Its efficacy is more pronounced when paired with ENNs, offering in some cases nearly $50\%$ better target domain accuracy when compared to CNNs trained without DA (\mytable{}~\ref{tab:classification_accuracies});
    \item We find that \algoname{} is effective across ENNs with varying group-order equivariance, and its performance improves as the degree of equivariance increases (\mytable{}~\ref{tab:grouporder} and \myfigure{}~\ref{fig:jsdistance});
    \item \algoname{} can improve the source and target domain performance of NNs, and its benefits are concretely seen when analyzing the clustering and alignment of NN latent spaces (\mytable{}~\ref{tab:silhouette} and \myfigure{}~\ref{fig:latent panel});
    \item Though not constructed as such, \algoname{} can inherently improve the calibration of trained NN-based classifiers (\mytable{}~\ref{tab:calibration_metrics});
    \item \algoname{} does not incur any considerable computational expense when training models, as it builds upon existing, efficient coding frameworks. 
    All models in this work were trained on one GPU in typically less than an hour.
\end{itemize}

There are multiple opportunities for further development of \algoname{}. 
First, the metric used for adjusting the dynamic Sinkhorn plan $S_\sigma$ relies on the pairwise norm between entries in the latent space. 
Other notions of distance to adjust the regularization of $S_\sigma$ can be considered. 
Second, in this work, we implemented a manual clipping of the $\eta_i$ terms in the loss function (Equation \ref{eqn:DOT}) to ensure that the DA loss does not overpower the classification loss. 
Other levels of clipping or regularization of the loss should be explored to determine the optimal balance between the two loss terms. 
Third, all experiments performed here utilize a fixed architecture that incorporates several features of NNs now considered standard, including dropout, pooling, and batch normalization. 
Notably missing are residual connections, which have been found to aid the convergence of many NNs. 
It would be interesting to study the efficacy of \algoname{} with deeper, more complex networks, which could be particularly beneficial in cases of severe domain shift, such as in the MRSSC2 dataset, as well as ENNs equivariant to the continuous analog of groups studied here (i.e., $\text{O}(2)$). 
Lastly, our method works with fixed classes and cannot operate when the classes between the source and target domains are not the same.
A potential extension of \algoname{} could involve making it compatible with a flexible number of classes in both the source and target domains, drawing inspiration from the DeepAstroUDA method \citep{deepastrouda}.

For more extreme covariate shifts, a multi-layer approach to \algoname{} as described in \mysection~\ref{sec:cross-mod} could result in better domain alignment. 
In this setting, one could impose domain alignment after intermediate convolutional layers with a similar dynamic scaling of $\sigma_\ell$ as used here. 
These can then be aggregated into a single DA loss term, or be treated separately as additional loss terms $\mathcal{L}_i$ paired with trainable coefficients $\eta_i$. 
This is a promising area of future work.

The problem of generalization in classification tasks in NNs can be primarily considered a problem of robust feature learning (e.g., architectural choice) and domain alignment.
The most successful domain adaptation methods should leverage principled choices for both aspects.
ENNs are natural candidates for robust feature learning because their feature learning capabilities can be inherently constrained to symmetries of the data.
However, most existing methods for domain adaptation implicitly require many empirical choices or hyperparameter tuning.
In this work, we have combined these aspects in introducing \algoname{}, which leverages a dynamic parameterization for OT-based DA hyperparameters during training, and works particularly well when paired with ENNs.
Our future work will be in refining this approach and further developing more automated DA algorithms.


\section*{Funding}
\label{sec:app:funding}

We acknowledge the Deep Skies Lab as a community of multi-domain experts and collaborators who’ve facilitated an environment of open discussion, idea generation, and collaboration. This community was important for the development of this project.


Notice: This work was produced by Fermi Forward Discovery Group, LLC under Contract No. 89243024CSC000002 with the U.S. Department of Energy, Office of Science, Office of High Energy Physics. The United States Government retains and the publisher, by accepting the work for publication, acknowledges that the United States Government retains a non-exclusive, paid-up, irrevocable, world-wide license to publish or reproduce the published form of this work, or allow others to do so, for United States Government purposes. The Department of Energy will provide public access to these results of federally sponsored research in accordance with the DOE Public Access Plan (\url{http://energy.gov/downloads/doe-public-access-plan}).

This work was also supported by the Office of Workforce Development for Teachers and Scientists, the Office of Science Graduate Student Research (SCGSR) program, and the National Science Foundation under Cooperative Agreement PHY-2019786 (The NSF AI Institute for Artificial Intelligence and Fundamental Interactions). The SCGSR program is administered by the Oak Ridge Institute for Science and Education for the DOE under contract number DE‐SC0014664. The Dunlap Institute is funded through an endowment established by the David Dunlap family and the University of Toronto. The U.S. Government retains and the publisher, by accepting the article for publication, acknowledges that the U.S. Government retains a non-exclusive, paid-up, irrevocable, world-wide license to publish or reproduce the published form of this manuscript, or allow others to do so, for U.S. Government purposes.

\section*{Acknowledgments and Author Contributions}
\label{sec:app:contributions}

The computations in this paper were run on the FASRC Cannon cluster, supported by the FAS Division of Science Research Computing Group at Harvard University. \\ 

\noindent The following authors contributed in different ways to the manuscript. \\

\noindent Pandya: Conceptualization, Methodology, Investigation, Formal analysis, Software, Validation, Data curation, Resources, Visualization, Writing - Original Draft, Writing - Review \& Editing, Project administration, Funding acquisition.\\ 

\noindent Patel: Methodology, Writing - Original Draft, Writing - Review \& Editing.\\ 

\noindent Nord: Methodology, Writing - Original Draft, Writing - Review \& Editing.\\ 

\noindent Walmsley: Data Curation, Writing - Review \& Editing \\

\noindent \'{C}iprijanovi\'{c}: Conceptualization, Methodology, Resources, Writing - Original Draft, Writing - Review \& Editing, Supervision, Project administration, Funding acquisition.\\

\bibliography{main}
\bibliographystyle{plainnat}


\newpage
\appendix
\onecolumn

\section{Training and Model Configurations}\label{app:networks}

\begin{table*}[!ht]
\centering
\caption{Model training configurations for each dataset. 
The learning rate followed a multiplicative decay schedule from $10^{-2}$ to $10^{-4}$, applied at epochs $\lfloor \text{Total Epochs} / 3 \rfloor$, and the best-performing model was selected based on validation loss.}\label{tab:training_configs}
\begin{tabular}{@{}l|l|c|c|c|c|c@{}}
\toprule
\textbf{Dataset} & \textbf{Model} & \textbf{Batch Size} & \textbf{Learning Rate} & \textbf{Optimizer} & \textbf{Warmup} & \textbf{Total Epochs} \\ \midrule

\multirow{4}{*}{Shapes} & CNN       & 128 & $10^{-2} \rightarrow 10^{-4}$ & AdamW & N/A  & 50 \\
                        & CNN-DA    & 128 & $10^{-2} \rightarrow 10^{-4}$ & AdamW & 10   & 50 \\
                        & $D_4$     & 128 & $10^{-2} \rightarrow 10^{-4}$ & AdamW & N/A  & 50 \\
                        & $D_4$-DA  & 128 & $10^{-2} \rightarrow 10^{-4}$ & AdamW & 5    & 50 \\ \midrule
                        
\multirow{4}{*}{\shortstack{Astro. \\ Objects}} & CNN       & 128 & $10^{-2} \rightarrow 10^{-4}$ & AdamW & N/A  & 50 \\
                        & CNN-DA    & 128 & $10^{-2} \rightarrow 10^{-4}$ & AdamW & 10   & 50 \\ 
                        & $D_4$     & 128 & $10^{-2} \rightarrow 10^{-4}$ & AdamW & N/A  & 50 \\
                        & $D_4$-DA  & 128 & $10^{-2} \rightarrow 10^{-4}$ & AdamW & 5    & 50 \\ \midrule
                        
\multirow{4}{*}{\shortstack{MNIST-M \\ (Noise)}} & CNN       & 128 & $10^{-2} \rightarrow 10^{-4}$ & AdamW & N/A  & 100 \\
                        & CNN-DA    & 128 & $10^{-2} \rightarrow 10^{-4}$ & AdamW & 30   & 100 \\ 
                        & $D_4$     & 128 & $10^{-2} \rightarrow 10^{-4}$ & AdamW & N/A  & 100 \\
                        & $D_4$-DA  & 128 & $10^{-2} \rightarrow 10^{-4}$ & AdamW & 20   & 100 \\ \midrule
                        
\multirow{4}{*}{\shortstack{MNIST-M \\ (PSF)}} & CNN       & 128 & $10^{-2} \rightarrow 10^{-4}$ & AdamW & N/A  & 100 \\
                        & CNN-DA    & 128 & $10^{-2} \rightarrow 10^{-4}$ & AdamW & 30   & 100 \\ 
                        & $D_4$     & 128 & $10^{-2} \rightarrow 10^{-4}$ & AdamW & N/A  & 100 \\
                        & $D_4$-DA  & 128 & $10^{-2} \rightarrow 10^{-4}$ & AdamW & 20   & 100 \\ \midrule
                        
\multirow{4}{*}{\shortstack{Galaxy Zoo \\ Evo}} & CNN       & 128 & $10^{-2} \rightarrow 10^{-4}$ & AdamW & N/A  & 100 \\
                        & CNN-DA    & 128 & $10^{-2} \rightarrow 10^{-4}$ & AdamW & 30   & 100 \\ 
                        & $D_4$     & 128 & $10^{-2} \rightarrow 10^{-4}$ & AdamW & N/A  & 100 \\
                        & $D_4$-DA  & 128 & $10^{-2} \rightarrow 10^{-4}$ & AdamW & 20   & 100 \\ \midrule

\multirow{4}{*}{MRSSC2} & CNN       & 128 & $10^{-2} \rightarrow 10^{-4}$ & AdamW & N/A  & 100 \\
                        & CNN-DA    & 128 & $10^{-2} \rightarrow 10^{-4}$ & AdamW & 30   & 100 \\ 
                        & $D_4$     & 128 & $10^{-2} \rightarrow 10^{-4}$ & AdamW & N/A  & 100 \\
                        & $D_4$-DA  & 128 & $10^{-2} \rightarrow 10^{-4}$ & AdamW & 20   & 100 \\ 

\bottomrule
\end{tabular}
\end{table*}

\begin{table}[ht]
    \centering
        \caption{CNN architecture used with the GZ Evo dataset. 
    For other experiments, the architecture only differs in the input dimension, dimension matching after convolutional layers, and output dimension (logits) size.}
    \begin{tabular}{|ccccc|}
        \hline
        \textbf{Layers} & \textbf{Properties} & \textbf{Stride} & \textbf{Padding} & \textbf{Output Shape} \\
        \hline
        Input & 3 x 100 x 100 & & & \\
        \hline
        Conv2D & Filters: 8 & 1 & 2 & (8, 100, 100) \\
        (w/ BatchNorm2D) & Kernel: 5x5 & & & \\
        & Activation: ReLU & & & \\
        \hline
        MaxPool2D & Kernel: 2x2 & 2 & 0 & (8, 50, 50) \\
        \hline
        Dropout & p=0.2 & & & (8, 50, 50) \\
        \hline
        Conv2D & Filters: 16 & 1 & 1 & (16, 50, 50) \\
        (w/ BatchNorm2D) & Kernel: 3x3 & & & \\
        & Activation: ReLU & & & \\
        \hline
        MaxPool2D & Kernel: 2x2 & 2 & 0 & (16, 25, 25) \\
        \hline
        Dropout & p=0.2 & & & (16, 25, 25) \\
        \hline
        Conv2D & Filters: 32 & 1 & 1 & (32, 25, 25) \\
        (w/ BatchNorm2D) & Kernel: 3x3 & & & \\
        & Activation: ReLU & & & \\
        \hline
        MaxPool2D & Kernel: 2x2 & 2 & 0 & (32, 12, 12) \\
        \hline
        Dropout & p=0.2 & & & (32, 12, 12) \\
        \hline
        Linear & Input Dimension: 4608 & & & (256) \\
        (w/ LayerNorm) & Output Dimension: 256 & & & \\
        & Activation: None & & & \\
        \hline
        Linear & Input Dimension: 256 & & & (6) \\
        & Output Dimension: 6 & & & \\
        & Activation: None & & & \\
        \hline
    \end{tabular}
    \label{table:convnet_galaxy}
\end{table}

\begin{table}[ht!]
    \centering
        \caption{
    $D_4$ ENN architecture used with the GZ Evo dataset. 
    For other experiments, the architecture only differs in the input dimension, dimension matching after convolutional layers, and output dimension (logits) size.}
    \begin{tabular}{|ccccc|}
        \hline
        \textbf{Layers} & \textbf{Properties} & \textbf{Stride} & \textbf{Padding} & \textbf{Output Shape} \\
        \hline
        Input & 3 x 100 x 100 & & & \\
        \hline
        R2Conv & Filters: 64 & 1 & 2 & (64, 100, 100) \\
        (w/ InnerBatchNorm) & Kernel: 5x5 & & & \\
        & Activation: ReLU & & & \\
        \hline
        PointwiseMaxPool2D & Kernel: 2x2 & 2 & 0 & (64, 50, 50) \\
        \hline
        PointwiseDropout & p=0.2 & & & (64, 50, 50) \\
        \hline
        R2Conv & Filters: 128 & 1 & 1 & (128, 50, 50) \\
        (w/ InnerBatchNorm) & Kernel: 3x3 & & & \\
        & Activation: ReLU & & & \\
        \hline
        PointwiseMaxPool2D & Kernel: 2x2 & 2 & 0 & (128, 25, 25) \\
        \hline
        PointwiseDropout & p=0.2 & & & (128, 25, 25) \\
        \hline
        R2Conv & Filters: 256 & 1 & 1 & (256, 25, 25) \\
        (w/ InnerBatchNorm) & Kernel: 3x3 & & & \\
        & Activation: ReLU & & & \\
        \hline
        PointwiseMaxPool2D & Kernel: 2x2 & 2 & 0 & (256, 12, 12) \\
        \hline
        PointwiseDropout & p=0.2 & & & (256, 12, 12) \\
        \hline
        GroupPooling & & & & (32, 12, 12) \\
        \hline
        Linear & Input Dimension: 4608 & & & (256) \\
        (w/ LayerNorm) & Output Dimension: 256 & & & \\
        & Activation: None & & & \\
        \hline
        Linear & Input Dimension: 256 & & & (6) \\
        & Output Dimension: 6 & & & \\
        & Activation: None & & & \\
        \hline
    \end{tabular}
\label{table:enn_galaxy_d4}
\end{table}

\end{document}